\pdfoutput=1
\documentclass[11pt]{article}
\usepackage[final]{acl}

\usepackage[table,xcdraw]{xcolor} 
\usepackage{times}
\usepackage{url}
\usepackage{latexsym}
\usepackage[T1]{fontenc}
\usepackage{sourcesanspro}
\usepackage[utf8]{inputenc}
\usepackage{xspace,microtype}
\usepackage{array}  
\usepackage{graphicx}
\usepackage{tcolorbox}
\usepackage{arydshln}
\usepackage{comment}
\usepackage{subcaption}
\usepackage{placeins}

\usepackage{dsfont}
\usepackage{amsmath}
\usepackage{amsfonts}
\usepackage{amssymb}
\usepackage{amsthm}
\usepackage{mathtools}
\usepackage{bm}
\usepackage[linesnumbered,ruled,vlined]{algorithm2e}
\SetAlgoCaptionLayout{small}
\usepackage{listings}
\usepackage{CJKutf8}

\usepackage{amssymb}
\newcommand{\cmark}{\textcolor{green!60!black}{\checkmark}}
\newcommand{\xmark}{\textcolor{red!70!black}{$\times$}}
\newcommand{\goldstar}{\textcolor{orange!80!black}{$\bigstar$}}

\usepackage{adjustbox}

\usepackage{enumitem}

\usepackage{tikz}
\usetikzlibrary{calc}
\usepackage[hang,flushmargin]{footmisc}

\usepackage{collcell}
\usepackage{booktabs}
\usepackage{multirow}
\definecolor{cK}{RGB}{221,221,221}
\definecolor{cR}{RGB}{255,204,204}
\definecolor{cG}{RGB}{204,255,204}
\definecolor{cB}{RGB}{204,204,255}
\definecolor{navyblue}{rgb}{0.0,0.0,0.5}
\definecolor{darkred}{rgb}{0.65, 0, 0}
\definecolor{darkgreen}{rgb}{0, 0.5, 0}
\definecolor{lB}{rgb}{0,0.4,0.9}
\definecolor{verylightblue}{RGB}{230,240,255}
\definecolor{lightblue}{RGB}{200,220,255}
\definecolor{mediumblue}{RGB}{150,190,255}
\definecolor{diaghl}{RGB}{255,235,205}
\definecolor{sig10}{RGB}{255,245,220} 
\definecolor{sig05}{RGB}{255,225,180}
\definecolor{sig01}{RGB}{255,180,100}
\definecolor{basecolor}{HTML}{EFEFEF} 
\definecolor{optimercolor}{HTML}{D9EAD3}

\definecolor{high1}{HTML}{DAEAF6}  
\definecolor{high2}{HTML}{FFF2CC}  
\definecolor{high3}{HTML}{D9EAD3}  
\definecolor{high4}{HTML}{F4CCCC}  
\definecolor{high5}{HTML}{E8E0F0}

\newcommand{\Sec}[1]{Section~\ref{sec:#1}}
\newcommand{\Fig}[1]{Figure~\ref{fig:#1}}
\newcommand{\Tab}[1]{Table~\ref{tab:#1}}
\newcommand{\Algo}[1]{Algorithm~\ref{algo:#1}}
\newcommand{\Eq}[1]{Eq.~(\ref{eq:#1})}

\newcommand{\formatrl}{\textls*[-20]{\textsc{Format}}\kern-0.05em\textsc{RL}}
\newcommand{\optimer}{\textsc{OptiMer}}

\usepackage{soul}
\sethlcolor{yellow}

\definecolor{myblue}{HTML}{2c7fb8}
\newcommand{\argmax}{\mathop{\rm arg~max}\limits}

\usepackage{hyperref}

\title{\optimer: Optimal Distribution Vector Merging\\Is Better than Data Mixing for Continual Pre-Training}

\author{
 \textbf{Haiyue Song} \textnormal{and} \textbf{Masao Utiyama}
\\
 National Institute of Information and Communications Technology, Kyoto, Japan\;
 \\
 \texttt{\{haiyue.song,mutiyama\}@nict.go.jp}\\
}

\begin{document}
\maketitle

\begin{abstract}
Continual pre-training is widely used to adapt LLMs to target languages and domains, yet the mixture ratio of training data remains a sensitive hyperparameter that is expensive to tune: they must be fixed \textit{before} training begins, and a suboptimal choice can waste weeks of compute.
In this work, we propose \textbf{\optimer}, which decouples ratio selection from training: we train one CPT model per dataset, extract each model's distribution vector, which represents the parameter shift induced by that dataset, and search for optimal composition weights \textit{post-hoc} via Bayesian optimization.
Experiments on Gemma 3 27B across languages (Japanese, Chinese) and domains (Math, Code) show that \optimer\ consistently outperforms data mixture and model averaging baselines with 15--35$\times$ lower search cost.
Key findings reveal that 1) the optimized weights can be interpreted as data mixture ratios, and retraining with these ratios improves data mixture CPT, and 2) the same vector pool can be re-optimized for a given objective without any retraining, producing target-tailored models on demand.
Our work establishes that data mixture ratio selection, traditionally a pre-training decision, can be reformulated as a post-hoc optimization over distribution vectors, offering a more flexible paradigm for continual pre-training.\footnote{Our code and model will be available at \url{https://github.com/shyyhs/optimer}.}
\end{abstract}

\section{Introduction}
\label{sec:intro}
Adapting large language models (LLMs) to specific languages and domains is a central challenge, driven by demand for both multilingual coverage and domain expertise~\cite{2504.05747,alnumay-etal-2025-command,2508.19667,2602.13033}.

Continual pre-training (CPT) is a common approach for such adaptation~\cite{gururangan-etal-2020-dont,ibrahim2024simple,yz2025investigating}, where the training corpus is typically a mixture of multiple datasets~\cite{fujii2024continual,sailor2report}.
However, the mixing ratio of these datasets is a critical yet sensitive hyperparameter: a suboptimal ratio can degrade model performance~\cite{xie2023doremi,ye2025data}.
Although recent methods estimate ratios via proxy models or small-scale experiments~\cite{xie2023doremi,liu2025regmix,ye2025data,2603.17945}, these estimates must be fixed before training begins and cannot be corrected afterward, meaning a poor choice may waste days or even weeks of GPU cluster time before its effect becomes apparent.
\begin{figure}[t]
  \centering
  \includegraphics[width=\linewidth]{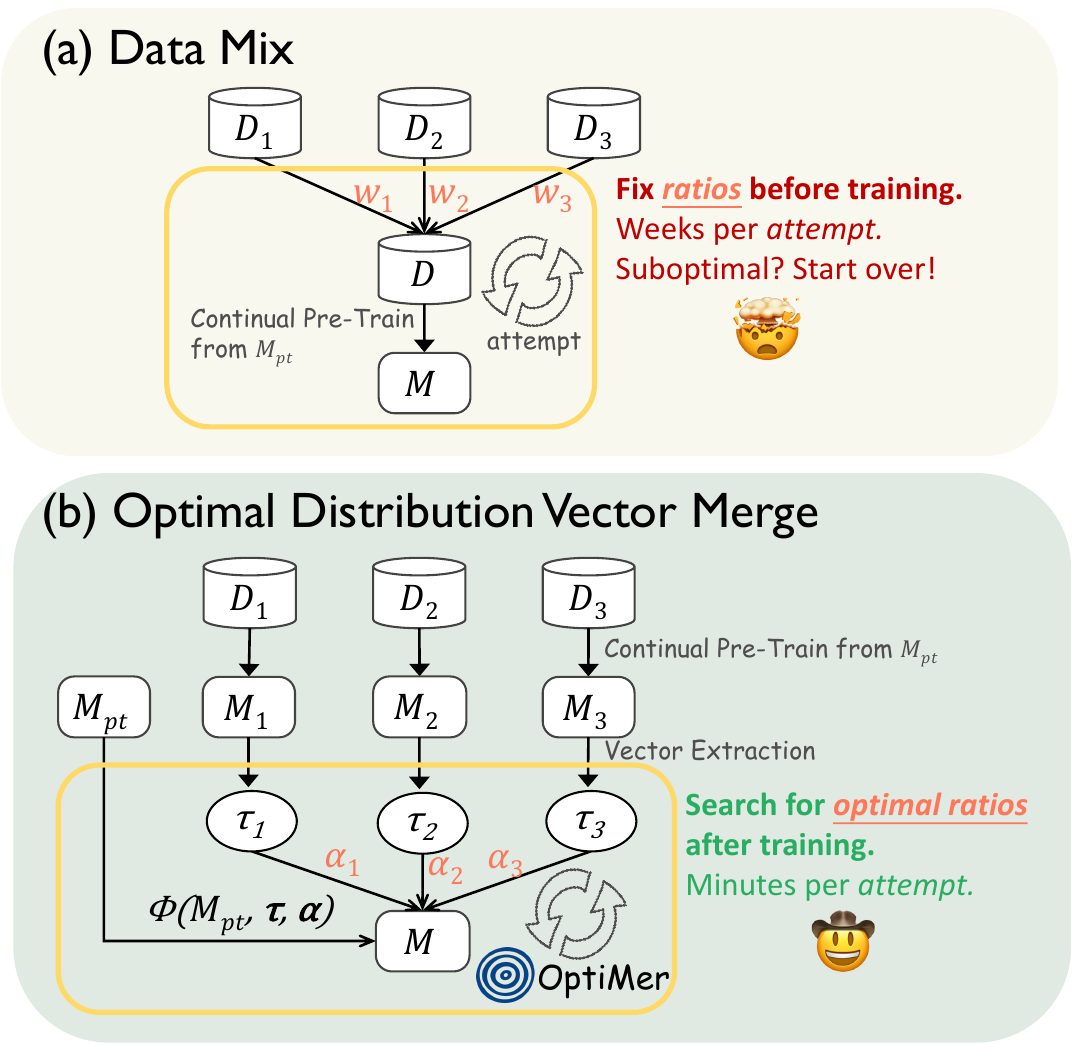}
  \caption{
    \textbf{Data Mix vs.\ \optimer.}
    (\textit{a})~Continual pre-training on a fixed data mixture requires
    the mixing ratios $\textcolor[HTML]{FD7758}{\{w_i\}}$ to be specified
    before training begins. Each attempt costs \textbf{days to weeks} of
    GPU time.
    (\textit{b})~Our approach trains one CPT model per dataset independently
    and extracts a distribution vector
    $\tau_i$ from each, which are then composed via a merge function
    $\Phi$ (e.g.\ DARE-Linear) with weights
    $\textcolor[HTML]{FD7758}{\{\alpha_i\}}$ optimized post-hoc.
    Each trial completes in \textbf{minutes}. The instruction-tuned vector is additionally merged in both settings.
  }
  \label{fig:motivation}
\end{figure}

To address this, we propose \optimer, which decouples data ratio selection from model training.
As illustrated in \Fig{motivation}, instead of fixing the data mixture ratio \textit{before training}, we train a separate CPT model on each dataset independently and extract the corresponding \textbf{distribution vector} (the parameter shift from the base PT model) \textit{after training}. 
Furthermore, rather than weight averaging, which leads to suboptimal performance~\cite{yadav2023tiesmerging}, \optimer\ searches for optimal merge weights via Bayesian optimization using the Tree-structured Parzen Estimator (TPE)~\cite{optuna_2019,watanabe2023tpe}. We find vector merging viable because vectors from distinct datasets are approximately orthogonal, allowing linear combination with minimal interference.

Experiments on Gemma 3 27B~\cite{2503.19786} with distribution vectors across languages (Japanese, Chinese) and domains (Math, Code) show that \optimer\ consistently outperforms data mixture baselines across all dataset combinations, while requiring 15--35$\times$ lower search time.
Moreover, the same collection of distribution vectors can be re-optimized toward different objectives, yielding multiple target-tailored models without any retraining.
Our contributions are as follows:
\begin{itemize}
    \setlength{\itemsep}{-1pt}
    \item We introduce the concept of \textbf{distribution vectors} for CPT and propose \textbf{\optimer}, a post-hoc framework that decouples data ratio selection from model training by optimizing merge weights via Bayesian optimization.
    \item Experimental results on 16 benchmarks covering five task groups (English, Japanese, Chinese, Math, Code) show that \optimer\ outperforms data mixture CPT and four model merging methods across three dataset combinations with 15--35$\times$ lower search cost. It further enables objective-specific re-optimization from a single vector pool without any re-CPT.
    \item Our analysis reveals that distribution vectors are approximately orthogonal (cosine 0.03--0.31), enabling composition without severe interference. Training dynamics show that CPT trajectories are approximately linear in parameter space, linking merge weights to effective training duration. \optimer\ search dynamics illustrate the sharp nature of the optimization landscape, thus highlighting the necessity of efficient searching rather than grid search. Additionally, optimized weights can serve as interpretable data mixture ratios and can be negative to remove cross-distribution interference. 
\end{itemize}
 
\section{Related Work}
\label{sec:related}

\paragraph{Continual Pre-training.}
Adapting a pretrained LLM to new languages or domains via CPT is a well-studied area \citep{gururangan-etal-2020-dont,2401.03129}.
It has been applied to language adaptation~\cite{fujii2024continual,dou-etal-2024-sailor,sailor2report} and domain adaptation~\cite{azerbayev2024llemma,2402.19173,pmcllama}.
Data mixture ratio is an important hyperparameter that largely affects model performance~\cite{2401.03129,shi2024continual}, which motivates work on data mixture optimization.

\paragraph{Data Mixture Optimization.}
Recently, several methods have been proposed to optimize data mixture ratios.
DoReMi~\cite{xie2023doremi} uses distributionally robust optimization on a small proxy model to produce domain weights for a larger target model.
RegMix~\cite{liu2025regmix} trains many small models on diverse mixtures and fits a regression to predict optimal ratios.
\citet{ye2025data,2603.17945} propose a predictive framework that transfers optimal ratios across scales.
Despite these advances, such methods must fix the ratio before training. 
Instead, we propose to adjust ratios post-hoc which avoids retraining.

\paragraph{Task Vectors and Model Merging.}
\citet{ilharco2023editing} show that task vectors $\tau = \theta_{\text{ft}} - \theta_{\text{base}}$, the difference between a fine-tuned model $\theta_{\text{ft}}$ and its base model $\theta_{\text{base}}$, can be composed via linear arithmetic to add or remove task capabilities, with subsequent work improving merging quality through sign conflict resolution~\cite{yadav2023tiesmerging} and delta sparsification~\cite{yu2024language}.
Chat Vector~\cite{huang-etal-2024-chat} applies weight arithmetic to transfer instruction-following capability to a CPT-adapted model without additional fine-tuning. Task-specific CPT checkpoint and LoRA adapter merging has also proven effective for finance domain~\cite{2511.02451} and machine translation~\cite{cao-etal-2026-completely}.
These works focus on task-specific transfer rather than improving general capability across multiple distributions.
In contrast, our work extends distribution vector composition to the multi-distribution CPT setting and achieves general performance improvement.

\paragraph{Automatic Merge Weight Search.}
Several methods automate the search for merge ratios, including test-time entropy minimization over per-layer weights~\cite{AdaMerging_ICLR_2024}, evolutionary search~\cite{akiba2025evolutionary}, and minimizing output divergence between merged and fine-tuned models~\cite{touayouch-etal-2026-divmerge}. They have been applied to at most two to three models or small-scale models due to the computational cost of the high-dimensional search spaces or population-based iterations.
Most relevant to our work, DEM~\cite{ram-etal-2024-dem} applies grid search over merge weights for SFT task vectors, but the cost of grid search increases exponentially with the number of vectors.
Our proposed \optimer\ replaces grid search with Bayesian optimization via TPE, achieving substantially higher theoretical search efficiency.

\section{Methodology}
\label{sec:method}
We define the notation and introduce distribution vectors in~\Sec{prelim}. In~\Sec{optimer}, we present \optimer, an automatic merge weight optimization approach via Bayesian optimization.

\subsection{Preliminaries}
\label{sec:prelim}
\paragraph{Notation.}
Let $\theta_{\mathrm{pt}} \in \mathbb{R}^d$ denote the parameters of a pretrained base model and $\theta_{\mathrm{it}}$ its instruction-tuned version.
Given $n$ data distributions $\{D_i\}_{i=1}^n$, each represented by a dataset, continual pre-training on $D_i$ from $\theta_{\mathrm{pt}}$ yields a CPT model $\theta_{\mathrm{CPT}_i}$.
\paragraph{Task Vectors.}
\citet{ilharco2023editing} define a \textit{task vector} $\tau = \theta_{\mathrm{ft}} - \theta_{\mathrm{base}}$ to capture the parameter change induced by fine-tuning, and construct a merged model as $\theta_{\mathrm{merge}} = \theta_{\mathrm{base}} + \lambda \tau$, where $\lambda$ is a scalar weight. This has been shown effective for adding or removing capabilities in the fine-tuning setting.

\paragraph{Distribution Vectors.}
We extend task vectors to the CPT setting. We define the \textbf{distribution vector} for $D_i$ as:
\begin{equation}
  \tau_i = \theta_{\mathrm{CPT}_i} - \theta_{\mathrm{pt}},
  \label{eq:dist_vec}
\end{equation}
which encodes the parameter change induced by distribution $D_i$.
Similarly, we extract an \textbf{IT vector} $\tau_{\mathrm{it}} = \theta_{\mathrm{it}} - \theta_{\mathrm{pt}}$ from the instruction-tuned model. Since our CPT models are trained from $\theta_{\mathrm{pt}}$, they lack instruction-following capability, and adding $\tau_{\mathrm{it}}$ recovers this capability without additional supervised fine-tuning~\cite{huang-etal-2024-chat}.

\paragraph{Multi-Vector Composition.}
A merged model incorporating $n$ distributions and instruction-following capability is constructed as:
\begin{equation}
  \theta_{\mathrm{merge}} = \theta_{\mathrm{pt}} + \alpha_{\mathrm{it}} \cdot \tau_{\mathrm{it}} + \sum_{i=1}^{n} \alpha_i \cdot \tau_i,
  \label{eq:merge}
\end{equation}
where $\{\alpha_i\}_{i=1}^n$ and $\alpha_{\mathrm{it}}$ are scalar merge weights.
Uniform weighting ($\alpha_i = 1/n$) is a natural baseline but leads to suboptimal performance in practice, as different distributions contribute unequally to the target objective.
The central question is then how to find the optimal weights $\boldsymbol{\alpha}^* = (\alpha_{\mathrm{it}}, \alpha_1, \ldots, \alpha_n)$ efficiently.

\subsection{\optimer: An Automatic Merge Weight Optimization Algorithm}
\label{sec:optimer}

\paragraph{Problem Formulation.}
The merged model in \Eq{merge} is parameterized by the weight vector $\boldsymbol{\alpha} = (\alpha_{\mathrm{it}}, \alpha_1, \ldots, \alpha_n) \in \mathbb{R}^{n+1}$.
We formulate the weight search as the optimization problem:
\begin{equation}
  \boldsymbol{\alpha}^* = \argmax_{\boldsymbol{\alpha}} \; \mathcal{S}\bigl(\theta_{\mathrm{merge}}(\boldsymbol{\alpha}),\, \mathcal{D}_{\mathrm{dev}}\bigr),
  \label{eq:optim}
\end{equation}
where $\mathcal{S}$ is an evaluation score computed on a development set $\mathcal{D}_{\mathrm{dev}}$.
Since $\mathcal{S}$ is obtained by running discrete benchmark evaluations, it provides no gradient with respect to $\boldsymbol{\alpha}$, making this a black-box optimization problem. A straightforward approach is grid search~\cite{ram-etal-2024-dem}, but its cost is $O(G^{n+1})$ for $G$ grid points per dimension, which becomes impractical as the number of vectors grows.
\paragraph{Bayesian Optimization via TPE.}
We solve \Eq{optim} using the Tree-structured Parzen Estimator~\cite{NIPS2011_86e8f7ab}, a Bayesian optimization method implemented in Optuna~\cite{optuna_2019}.
Given $N$ observed trials, where each trial consists of constructing a merged model with a candidate $\boldsymbol{\alpha}$ and evaluating it, TPE partitions the $N$ trials by a quantile $\gamma$ (e.g. 10\%) into a \emph{good} set and a \emph{bad} set based on their performance on $\mathcal{D}_{\mathrm{dev}}$.
Two separate density models are estimated via kernel density estimation:
\begin{equation}
  p(\boldsymbol{\alpha} \mid \mathcal{S}) =
  \begin{cases}
    \ell(\boldsymbol{\alpha}) & \text{if } \mathcal{S} \geq s^*, \\
    g(\boldsymbol{\alpha}) & \text{if } \mathcal{S} < s^*,
  \end{cases}
  \label{eq:tpe}
\end{equation}
where $s^*$ is the top-$\gamma$ quantile of observed scores, $\ell$ models the density of high-scoring configurations, and $g$ models the rest.
The next candidate is selected by maximizing the ratio $\ell(\boldsymbol{\alpha}) / g(\boldsymbol{\alpha})$, concentrating sampling in promising regions of the weight space.
While grid search requires $O(G^{n+1})$ evaluations for $G$ grid points per dimension, TPE typically converges in $O(10n)$ trials~\cite{watanabe2023tpe}, making it practical even as the number of vectors grows.
Furthermore, TPE can sample candidates independently, enabling parallel trial execution on multiple GPUs.

\begin{algorithm}[t]
\DontPrintSemicolon
\SetAlgoLined
\KwIn{Base model $\theta_{\mathrm{pt}}$,
      distribution vectors $\{\tau_i\}_{i=1}^n$,
      IT vector $\tau_{\mathrm{it}}$,
      dev set $\mathcal{D}_{\mathrm{dev}}$,
      number of trials $T$, top-$K$}
\KwOut{Optimized weights $\boldsymbol{\alpha}^*$, merged model $\theta_{\mathrm{merge}}^*$}
Sample $N_0$ random trials to initialize density models $\ell$, $g$\;
\For{$t = N_0 + 1$ \KwTo $T$}{
  $\boldsymbol{\alpha}^{(t)} \leftarrow \argmax_{\boldsymbol{\alpha}}\; \ell(\boldsymbol{\alpha}) / g(\boldsymbol{\alpha})$\;
  $\theta^{(t)}_{\mathrm{merge}} \leftarrow
    \theta_{\mathrm{pt}}
    + \alpha_{\mathrm{it}}^{(t)}\,\tau_{\mathrm{it}}
    + \sum_{i=1}^{n} \alpha_i^{(t)}\,\tau_i$\;
  $s^{(t)} \leftarrow
    \mathcal{S}(\theta^{(t)}_{\mathrm{merge}},\,\mathcal{D}_{\mathrm{dev}})$\;
  Update $\ell$, $g$ with $(\boldsymbol{\alpha}^{(t)}, s^{(t)})$\;
}
Re-evaluate top-$K$ trials on full $\mathcal{D}_{\mathrm{dev}}$\;
$\boldsymbol{\alpha}^* \leftarrow \argmax_{\text{top-}K}\; \mathcal{S}$\;
$\theta_{\mathrm{merge}}^* \leftarrow \theta_{\mathrm{pt}} + \alpha_{\mathrm{it}}^*\,\tau_{\mathrm{it}} + \sum_{i=1}^{n} \alpha_i^*\,\tau_i$\;
\Return{$\boldsymbol{\alpha}^*,\; \theta_{\mathrm{merge}}^*$}
\caption{\optimer}
\label{algo:optuna}
\end{algorithm}

\paragraph{Algorithm.}
\Algo{optuna} summarizes the \optimer\ pipeline.
The search begins with $N_0$ random trials to initialize the TPE density models $\ell$ and $g$.
In subsequent trials, TPE proposes a candidate $\boldsymbol{\alpha}^{(t)}$ by maximizing $\ell / g$; then a merged model is constructed via \Eq{merge} and scored on a subset of the development set; finally the density models are updated with the new observation.
After $T$ trials, the top-$K$ configurations are re-evaluated on the full development set to obtain the final model $\theta_{\mathrm{merge}}^*$.

\section{Experimental Settings}
\label{sec:exp}
This section describes continued pre-training configuration (\S\ref{sec:cpt-settings}), merge settings and \optimer\ hyperparameters (\S\ref{sec:merge-settings}), baseline settings (\S\ref{sec:baselines}), and evaluation settings (\S\ref{sec:eval-settings}).

\subsection{Continual Pre-Training Settings}
\label{sec:cpt-settings}
We sampled training data from the LLM-jp Corpus v4~\cite{2407.03963}\footnote{\url{https://gitlab.llm-jp.nii.ac.jp/datasets/llm-jp-corpus-v4}} to construct CPT datasets across languages (Japanese, Chinese) and domains (Math, Code), each containing 1B tokens.
For data mixture baselines, datasets were combined at equal ratios with $n$B tokens in total.
We continually pre-trained \texttt{gemma-3-27b-pt}~\cite{2503.19786} for 1 epoch (2{,}000 steps) on each dataset, with sequences packed to 4{,}096 tokens and an effective batch size of 128.
Following \citet{fujii2024continual}, we use AdamW~\cite{loshchilov2018decoupled} ($\beta_1{=}0.90$, $\beta_2{=}0.95$, weight decay $0.1$, gradient clipping $1.0$) with a peak learning rate of $4{\times}10^{-5}$ and cosine decay to $1\%$. We report the effect of different hyperparameter settings in Appendix~\ref{sec:appendix-hparam}.
Training used BFloat16 with DeepSpeed ZeRO Stage~3 on 8 NVIDIA H200 GPUs (141\,GB) on the ABCI 3.0 cluster.

\subsection{Merge Settings}
\label{sec:merge-settings}

We used \texttt{gemma-3-27b-it}~\cite{2503.19786} to calculate $\tau_{\mathrm{it}}$.
Merges were performed with DARE-Linear~\cite{yu2024language} via \texttt{mergekit}~\cite{goddard-etal-2024-arcees}, excluding embedding and positional layers (\texttt{embed\_tokens}, \texttt{lm\_head}, \texttt{rotary}) to preserve the base model's token representations. 

\vspace{3pt}
\noindent
For \optimer, we ran $T{=}100$ trials with the TPE sampler ($N_0{=}20$ random startup trials), executed in parallel across 8 GPUs.
The search space was set to $\alpha_{\mathrm{it}} \in [0.3, 1.0]$ and $\alpha_{CPT_i} \in [0.0, 1.0]$.

\vspace{3pt}
\noindent
Proxy tasks were selected to match the target axes of each merge experiment (e.g., \texttt{gsm8k} and \texttt{ja\_leaderboard\_mgsm} for a Japanese$\times$Math merge).
During the search, each trial was scored on the first 100 samples per proxy task for efficiency.
The top-$K{=}3$ trials were then re-evaluated on the first 300 samples per task as the development set.

\begin{table*}[t]
\centering
\resizebox{\textwidth}{!}{%
\begin{tabular}{ll rrrr r rr r r r}
\toprule
& & \multicolumn{4}{c}{\textbf{English}} & \textbf{Math} & \multicolumn{2}{c}{\textbf{Code}} & \textbf{Japanese} & \textbf{Chinese} & \\
\cmidrule(lr){3-6}\cmidrule(lr){7-7}\cmidrule(lr){8-9}\cmidrule(lr){10-10}\cmidrule(lr){11-11}
\multirow{-2}{*}{\textbf{Combination}} & \multirow{-2}{*}{\textbf{Method}}
  & MMLU & ARC-C & HellaSwag & TQA
  & GSM8K
  & HE & MBPP
  & JA LB
  & C-Eval
  & \multirow{-2}{*}{\textbf{Avg.}} \\
\midrule
& Gemma 3 27B PT
  & 31.59 & 27.30 & 48.41 & 54.71
  & 0.83 & 0.00 & 0.60
  & 39.04 & 27.41 & 25.54 \\
\vspace{6pt}
& Gemma 3 27B IT
  & 51.03 & 34.90 & 55.91 & 64.87
  & 15.39 & 2.44 & 37.80
  & 54.64 & 37.96 & 39.44 \\
& \multicolumn{11}{l}{\textit{Single CPT, merged with IT ($\alpha_{\mathrm{it}}{=}0.6$)}} \\
& CPT$_{\text{En}}$ 
  & 73.60 & 69.80 & 82.82 & 46.39
  & 79.98 & 71.95 & 68.40
  & 43.06 & 62.56 & 66.51 \\
  & CPT$_{\text{Ja}}$
    & 72.11 & 67.06 & 81.98 & 55.32
    & 83.78 & 68.29 & 68.20
    & 72.50 & 62.48 & 70.19 \\
  & CPT$_{\text{Zh}}$
    & 72.85 & 68.94 & 81.70 & 48.84
    & 82.18 & 71.95 & 67.80
    & 57.98 & 62.85 & 68.34 \\
  & CPT$_{\text{Math}}$
    & 75.07 & 70.14 & 82.72 & 43.57
    & 82.64 & 57.32 & 68.00
    & 72.79 & 63.37 & 68.40 \\
  & CPT$_{\text{Code}}$
    & 71.86 & 70.05 & 81.71 & 51.16
    & 81.96 & 64.63 & 68.40
    & 73.39 & 62.26 & 69.49 \\
\midrule
\rowcolor{basecolor}
\multirow{6}{*}{\cellcolor{white}Ja + Math}
  \cellcolor{white} & DataMix$_{\alpha_{\mathrm{it}}=0.6}$ (Baseline)
    & 73.17 & 68.77 & 81.72 & 49.57
    & 83.62 & 50.00 & 68.80
    & 73.34 & 61.74 & 67.86 \\
& \textbf{DataMix$_{\text{OptiMer ratio}}$ (Ours)}
    & 72.72 & \textbf{70.05} & 82.62 & 50.06
    & 76.19 & \textbf{67.07} & 64.80
    & 72.85 & 61.96 & 68.70 \\
    \cmidrule(l){2-12}
  & \textit{Task Arithmetic}~\cite{ilharco2023editing}
    & \textbf{75.44} & 69.88 & 83.31 & 41.49
    & 74.45 & 60.37 & 1.60
    & \textbf{75.11} & \textbf{63.67} & 60.59 \\
  & \textit{TIES}~\cite{yadav2023tiesmerging}
    & 71.14 & 65.70 & 80.03 & 30.11
    & 69.52 & 26.22 & 0.00
    & 72.40 & 58.84 & 52.66 \\
  & \textit{DARE-Linear}~\cite{yu2024language}
    & 75.02 & 69.71 & \textbf{83.74} & 42.96
    & 78.17 & 58.54 & 65.00
    & 72.37 & 63.30 & 67.65 \\
  & \textit{DARE-TIES}~\cite{yu2024language}
    & 74.72 & 68.94 & 82.75 & 39.53
    & 73.77 & 60.98 & 0.80
    & 74.39 & 62.33 & 59.80 \\
    \rowcolor{optimercolor}
  \cellcolor{white}& \textbf{\optimer\ (Ours)}
    & 73.17 & 68.34 & 82.50 & \textbf{55.20}
    & \textbf{84.46} & 59.76 & \textbf{70.40}
    & 72.53 & 63.45 & \textbf{69.98} \\
\midrule
\rowcolor{basecolor}
\multirow{6}{*}{\cellcolor{white}Ja + Code}
  & DataMix$_{\alpha_{\mathrm{it}}=0.6}$ (Baseline)
    & 71.23 & 69.20 & 81.14 & 43.08
    & 78.62 & 64.02 & 65.00
    & 72.38 & 60.40 & 67.23 \\
& \textbf{DataMix$_{\text{OptiMer ratio}}$} (Ours)
    & \textbf{72.35} & 68.60 & 80.22 & 51.77
    & 74.00 & 55.49 & 67.60
    & \textbf{74.66} & \textbf{62.26} & 67.44 \\
     \cmidrule(l){2-12}
  & \textit{Task Arithmetic}~\cite{ilharco2023editing}
    & 72.33 & \textbf{70.39} & 82.44 & 42.72
    & 78.17 & 45.12 & 60.20
    & 74.04 & 61.96 & 65.26 \\
  & \textit{TIES}~\cite{yadav2023tiesmerging}
    & 66.44 & 66.98 & 79.69 & 36.11
    & 70.58 & 43.29 & 53.60
    & 70.22 & 54.75 & 60.18 \\
  & \textit{DARE-Linear}~\cite{yu2024language}
    & 72.33 & 70.31 & \textbf{82.50} & 42.84
    & 79.08 & 46.34 & 60.60
    & 74.07 & 62.18 & 65.58 \\
  & \textit{DARE-TIES}~\cite{yu2024language}
    & 71.14 & 69.37 & 82.00 & 41.49
    & 76.04 & 42.68 & 60.40
    & 73.30 & 60.25 & 64.07 \\
    \rowcolor{optimercolor}
  \cellcolor{white}& \textbf{\optimer\ (Ours)}
    & 71.41 & 66.89 & 81.44 & \textbf{54.22}
    & \textbf{83.47} & \textbf{64.63} & \textbf{70.20}
    & 73.06 & 61.81 & \textbf{69.68} \\
\midrule
\rowcolor{basecolor}
\multirow{6}{*}{\cellcolor{white}Ja + Zh + Math}
  & DataMix$_{\alpha_{\mathrm{it}}=0.6}$ (Baseline)
    & 73.83 & 69.03 & 80.54 & 37.70
    & 79.61 & 32.32 & 65.40
    & 72.77 & 62.18 & 63.71 \\
  & \textbf{DataMix$_{\text{OptiMer ratio}}$ (Ours)}
    & 74.34 & 67.58 & 81.80 & 44.80
    & 81.27 & 66.46 & 67.20
    & \textbf{74.05} & 60.48 & 68.66 \\
     \cmidrule(l){2-12}
  & \textit{Task Arithmetic}~\cite{ilharco2023editing}
    & \textbf{75.27} & \textbf{69.97} & 83.18 & 38.92
    & 73.54 & 45.73 & 0.40
    & 73.17 & \textbf{64.41} & 58.29 \\
  & \textit{TIES}~\cite{yadav2023tiesmerging}
    & 70.13 & 66.72 & 79.65 & 26.68
    & 65.81 & 14.63 & 0.20
    & 69.86 & 60.03 & 50.41 \\
  & \textit{DARE-Linear}~\cite{yu2024language}
    & 74.48 & 69.28 & \textbf{83.65} & 38.68
    & 76.42 & 50.00 & 62.00
    & 70.92 & 63.89 & 65.48 \\
  & \textit{DARE-TIES}~\cite{yu2024language}
    & 74.06 & 69.54 & 82.59 & 35.86
    & 70.36 & 45.12 & 0.40
    & 72.38 & 63.60 & 57.10 \\
    \rowcolor{optimercolor}
  \cellcolor{white}& \textbf{\optimer\ (Ours)}
    & 72.94 & 68.60 & 82.98 & \textbf{51.77}
    & \textbf{83.93} & \textbf{67.07} & \textbf{70.20}
    & 72.65 & 63.22 & \textbf{70.37} \\
\bottomrule
\end{tabular}%
}
\caption{%
  Main results comparing \optimer\ with baselines across dataset combinations.
  \textbf{Single}: individual domain CPT merged with IT ($\alpha_{\mathrm{it}}=0.6$);
   {\sethlcolor{basecolor}\hl{\textbf{DataMix}}}: multi-domain CPT with equal data ratios, merged with IT ($\alpha_{\mathrm{it}}=0.6$);
  {\sethlcolor{optimercolor}\hl{\textbf{\textsc{OptiMer}}}}: Bayesian-optimized merge weights over all distribution vectors. \textbf{Task Arithmetic}, \textbf{TIES}, and \textbf{DARE} apply uniform weighting among all models including the IT model.
  \textbf{Bold} indicates the best result per combination.
  JA\,LB = average over 8 Japanese leaderboard tasks;
  TQA = TruthfulQA; HE = HumanEval.
}
\label{tab:main}
\end{table*}

\subsection{Baselines}
\label{sec:baselines}
We compared with the following baseline methods.

\vspace{3pt}
\noindent \textbf{DataMix.}
We trained a single CPT model on the concatenation of all datasets in each combination ($n$B tokens in total for $n$ datasets), and merged it with the IT vector with optimized hyperparameter $\alpha_{\mathrm{it}}{=}0.6$ (\S\ref{sec:appendix-hparam}) to recover IT capability. 

\vspace{3pt}
\noindent \textbf{DataMix$_{\text{OptiMer ratio}}$} models were trained on the same $n$B tokens with the data mixing ratio directly derived from the optimal merge weights found by \optimer.

\vspace{3pt}
\noindent \textbf{Average Merge.}
We merged $n$ CPT vectors and the IT vector using equal weights via Task Arithmetic~\cite{ilharco2023editing}, TIES~\cite{yadav2023tiesmerging}, and DARE~\cite{yu2024language}.

\subsection{Evaluation Settings}
\label{sec:eval-settings}
We used the \texttt{lm-evaluation-harness}~\cite{eval-harness} framework with the vLLM backend~\cite{kwon2023efficient}, \textit{using 1-shot prompting across all tasks} on these five task groups:
\textit{En} (MMLU~\cite{hendrycks2021measuring}, ARC-Challenge~\cite{allenai:arc}, HellaSwag~\cite{zellers-etal-2019-hellaswag}, TruthfulQA~\cite{lin-etal-2022-truthfulqa}),
\textit{Ja} (8 tasks from the Japanese Leaderboard~\cite{leaderboard_mgsm,ja_leaderboard_xlsum,jglue_kurihara-etal-2022-jglue,xwinograd_tikhonov2021heads}
\footnote{\url{https://github.com/EleutherAI/lm-evaluation-harness/blob/main/lm_eval/tasks/japanese_leaderboard/README.md}})
,
\textit{Zh} (C-Eval~\cite{huang2023ceval}),
\textit{Math} (GSM8K~\cite{leaderboard_mgsm}),
and \textit{Code} (HumanEval~\cite{2107.03374}, MBPP~\cite{2108.07732}). Detailed descriptions of each benchmark are provided in Appendix~\ref{sec:benchmark_details}.
We also report \textbf{Avg.}, which is the unweighted mean of all tasks, with the Japanese Leaderboard benchmark calculated as one task.

\section{Results and Analysis}
\label{sec:results}
We compare \optimer\ to baselines (\S\ref{sec:main-results}), analyze distribution vectors (\S\ref{sec:vector-analysis}), training dynamics (\S\ref{sec:cpt-dynamics}), and optimization dynamics (\S\ref{sec:optimer-dynamics}) to understand how and why \optimer\ works. We further conduct experiments with negative vector weight (\S\ref{sec:weight-analysis}).
Finally, we apply \optimer\ to build a Japanese-optimized LLM (\S\ref{sec:sealion}).

\begin{table*}[t]
\centering
\resizebox{\textwidth}{!}{%
\begin{tabular}{l rrrr r rr r r r}
\toprule
& \multicolumn{4}{c}{\textbf{English}} & \textbf{Math} & \multicolumn{2}{c}{\textbf{Code}} & \textbf{Japanese} & \textbf{Chinese} & \\
\cmidrule(lr){2-5}\cmidrule(lr){6-6}\cmidrule(lr){7-8}\cmidrule(lr){9-9}\cmidrule(lr){10-10}
\multirow{-2}{*}{\textbf{Objective}}
  & MMLU & ARC-C & HellaSwag & TQA
  & GSM8K
  & HE & MBPP
  & JA LB
  & C-Eval
  & \multirow{-2}{*}{\textbf{Avg.}} \\
\midrule
English
  & \cellcolor{high1}72.87          & \cellcolor{high1}\textbf{69.62} & \cellcolor{high1}82.02          & \cellcolor{high1}\textbf{47.74}
  & 82.49          & 44.51          & 71.40
  & 69.56          & 62.93          & 65.35 \\
Math
  & 70.10          & 63.05          & 79.16          & 41.13
  & \cellcolor{high2}82.87 & 4.88           & \textbf{72.60}
  & 68.47          & 58.99          & 58.37 \\
Japanese
  & \textbf{73.96} & 68.86          & \textbf{82.73} & 46.39
  & \textbf{84.69} & \textbf{68.29} & 69.20
  & \cellcolor{high3}\textbf{73.37} & 62.85          & \textbf{69.07} \\
Chinese
  & 71.93          & 69.11          & 82.37          & 34.76
  & 72.71          & 11.59          & 1.60
  & 40.78          & \cellcolor{high4}\textbf{63.60} & 44.29 \\
\bottomrule
\end{tabular}%
}
\caption{%
  Flexibility of \optimer: the same vector pool
  \{En, Math, Ja, Zh\} is re-optimized for one of the four task groups \{{\sethlcolor{high1}\hl{En}}, {\sethlcolor{high2}\hl{Math}}, {\sethlcolor{high3}\hl{Ja}}, {\sethlcolor{high4}\hl{Zh}}\} in four separate runs, each with 100 trials.
  \textbf{Bold} indicates the best value in each column.
  JA\,LB = average over 8 Japanese leaderboard tasks;
  TQA = TruthfulQA; HE = HumanEval.
}
\label{tab:flexibility}
\end{table*}

\subsection{Main Results}
\label{sec:main-results}

\paragraph{Performance.}
As shown in \Tab{main}, \optimer\ achieves the highest average score across all dataset combinations, outperforming the DataMix baseline in each group by 2.1--6.7 points. We make the following observations:
(i) Single-domain CPT models already perform well, yet DataMix shows lower performance despite using more training data, indicating its sensitivity to suboptimal mixture ratios.
(ii) Model averaging methods such as DARE-Linear achieve reasonable overall scores, but suffer from catastrophic failures on Code tasks. After inspecting outputs, we found these models generate syntactically malformed code (e.g., missing indentation), rather than hallucinated content.
(iii) \optimer\ maintains strong TruthfulQA (TQA) scores (51--55) where all other methods degrade significantly (30--49), suggesting that optimized weights better preserve the base model's calibration. We present case studies in Appendix~\ref{sec:case_study} to illustrate their qualitative difference.

Additionally, optimal merge weights can be interpreted as post-hoc data mixture ratios. \textbf{DataMix$_{\text{OptiMer ratio}}$} first converts weights into dataset proportions and retrains DataMix models with these ratios to form a training set with 2B or 3B data. 
Across all combinations, it outperforms the uniform ratio DataMix baselines, e.g., in Ja+Zh+Math, the average improves from 63.71 to 68.66.
This confirms DataMix suffers from suboptimal ratio selection, and \optimer\ discovers better ratios without further training.
Furthermore, \optimer\ still achieves the best performance, suggesting the advantage of post-hoc composition.

\begin{figure}[t]
    \centering
    \includegraphics[width=\columnwidth]{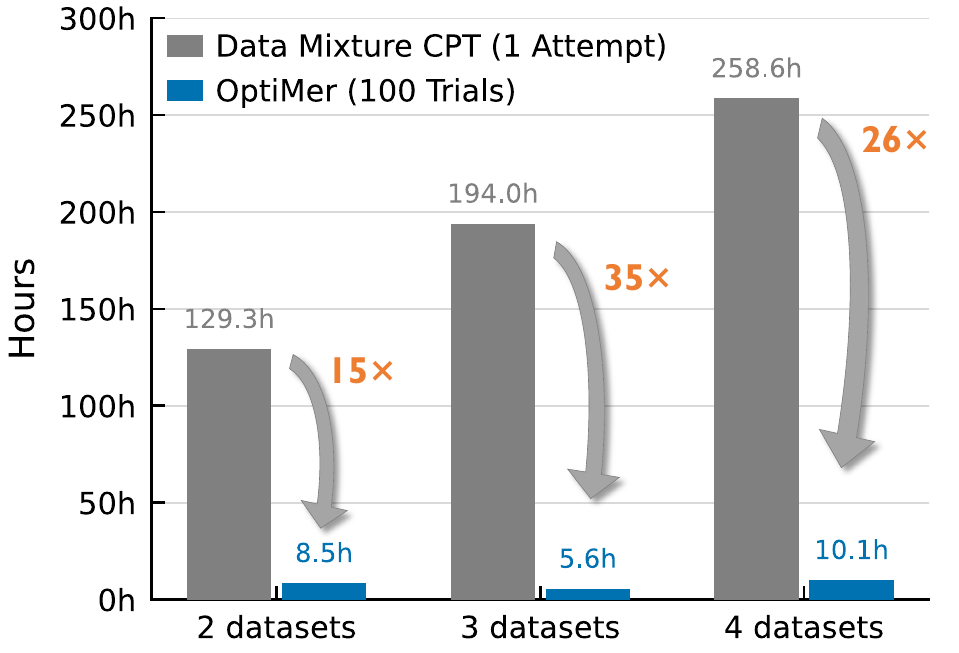}
    \caption{\textbf{Computational cost} comparison between data mixture CPT and \optimer\ across different numbers of datasets \textbf{during ratio optimization}. Training cost excluded as it is identical for both approaches.}
    \label{fig:efficiency}
\end{figure}

\paragraph{Efficiency.}
\optimer\ is 15--35$\times$ faster than DataMix for searching optimal ratios, and this advantage becomes larger with more datasets, as shown in \Fig{efficiency}.
In ratio searching, a 100-trial \optimer\ search completes in 8.6 hours, compared to 128.9 hours for a single DataMix run.
We found each \optimer\ trial consists of a merge (10.2\% of trial time) and an evaluation (89.8\%), so the cost is nearly constant regardless of $n$, whereas DataMix cost scales with the data size.
Note \textbf{that the training cost is the same}: \optimer\ trains $n$ models on 1B tokens each, while DataMix trains one model on $n$B tokens.


\paragraph{Flexibility.}
\optimer\ can produce an objective-optimized model on demand without retraining. \Tab{flexibility} shows the results re-optimizing for different objectives using the same four distribution vectors \{Ja, Zh, En, Math\}. We found that (i) in most cases, the model optimized for a given objective yields the highest score on its target tasks (e.g., the model optimized for Chinese tasks achieves the best C-Eval score), and  (ii) the Japanese-optimized model also achieves the best overall performance, suggesting that Japanese data also benefits multilingual performance. We leave the investigation of this cross-lingual transfer effect to future work.
\begin{figure}[t]
    \centering
    \includegraphics[width=\columnwidth]{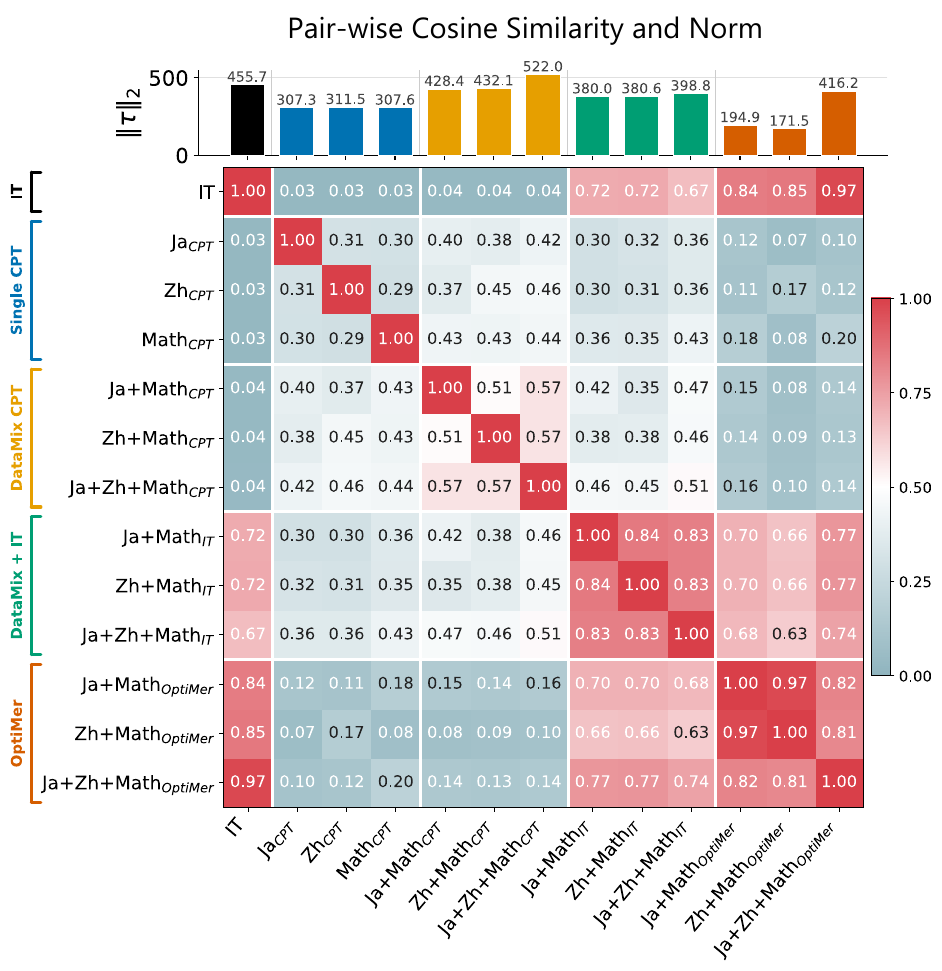}
    \caption{Pairwise cosine similarity and norm of distribution vectors.}
    \label{fig:heatmap}
\end{figure}
\begin{figure}[t]
    \centering
    \includegraphics[width=\columnwidth]{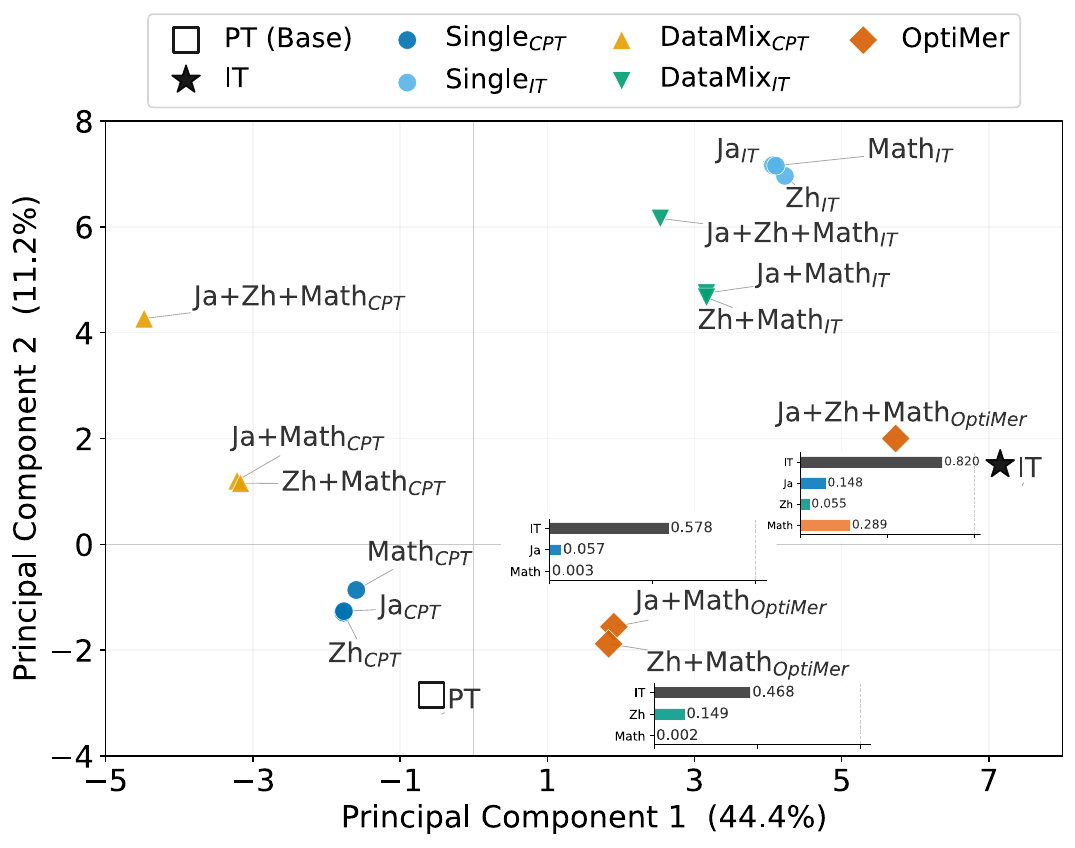}
    \caption{PCA projection of distribution vectors with \optimer\ merge weights in bar charts.}
    \label{fig:pca}
\end{figure}

\subsection{Analysis of Distribution Vectors}
\label{sec:vector-analysis}
We show the pair-wise cosine similarity of distribution vectors (i.e. $\theta - \theta_{\mathrm{pt}}$) in \Fig{heatmap}. We found CPT and IT vectors are nearly orthogonal (cosine $\approx$ 0.03), and different CPT vectors also exhibited low similarity (0.29--0.31), indicating that each distribution modifies an independent subspace, which supports the feasibility of linear composition. 
Layer-wise similarity analysis and cosine similarity of more models are shown in Appendix~\ref{sec:heatmap_full}.

\noindent \Fig{pca} visualizes the same vectors (sparsified through layer-wise truncated SVD) via PCA. The accompanying bar charts show the optimal merge weights. Both confirm that CPT vectors lie far from the IT vector.
Two additional insights emerge from combining both figures:
(i) DataMix models with more datasets drift further from IT (in both cosine similarity and PCA distance), showing that CPT dilutes IT capability, while \optimer\ is unaffected, maintaining a cosine similarity of greater than 0.97 with $\theta_{it}$. This explains the widening performance gap in \Tab{main} (+2.1 for 2-way vs.\ +6.7 for 3-way).
(ii) \optimer\ assigns large weight to IT and small weights to CPT vectors, suggesting that IT-targeted perturbation is more effective than uniform averaging.

\begin{figure}[th]
    \centering
    \includegraphics[width=\columnwidth]{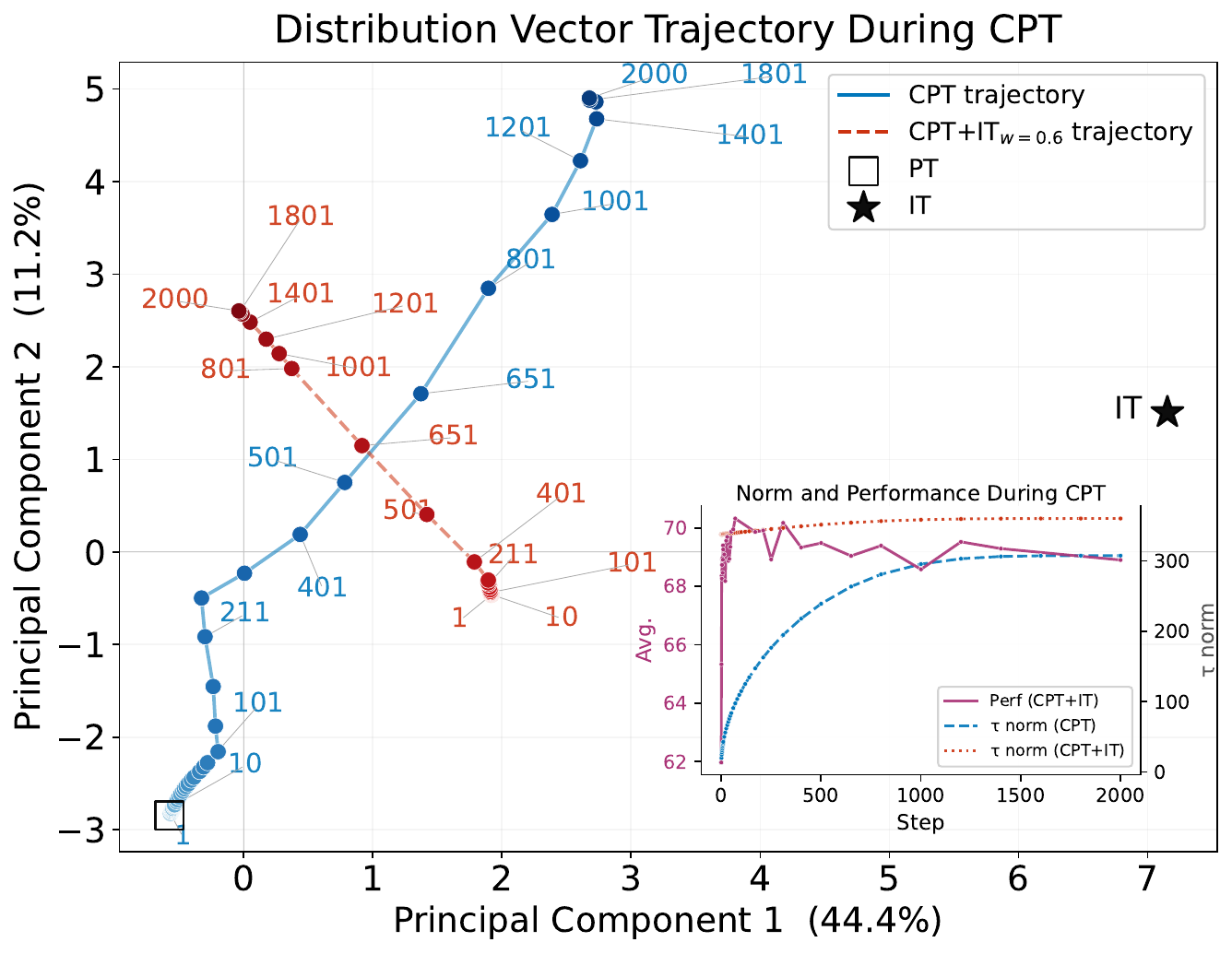}
    \caption{Distribution vector trajectories during CPT on 1B Japanese data, projected onto the same PCA space as used in \Fig{pca}.}
    \label{fig:train_steps}
\end{figure}

\subsection{Continual Pre-Training Dynamics}
\label{sec:cpt-dynamics}
We show the vector trajectories during training in~\Fig{train_steps}. We observed that both the \textit{CPT} and \textit{CPT merged with IT} vector trajectories move away from the IT vector, with rapid change in early steps. We found performance reacheed peak in the early stage with small vector norm, which is consistent with the thicket regime phenomenon~\cite{2603.12228}, and decreased gradually, possibly due to the divergence from the base model~\cite{2601.09473}.
Furthermore, the CPT trajectory is approximately linear, indicating that adjusting the merge weight $\alpha_i$ is analogous to controlling the effective training duration, which also explains why \optimer\ assigns small CPT weights.


\subsection{OptiMer Search Dynamics}
\label{sec:optimer-dynamics}
\begin{figure*}[!htbp]
    \centering
    \begin{subfigure}[t]{0.48\textwidth}
        \centering
        \includegraphics[width=\textwidth]{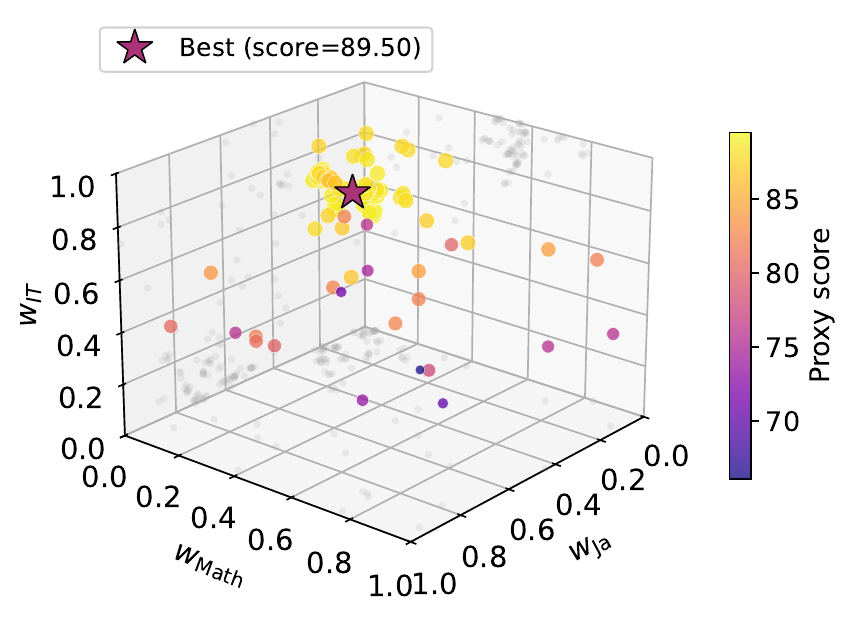}
        \caption{Exploration of the merge weight space. Each point represents one trial; color indicates proxy score.}
        \label{fig:3d}
    \end{subfigure}
    \hfill
    \begin{subfigure}[t]{0.48\textwidth}
        \centering
        \includegraphics[width=\textwidth]{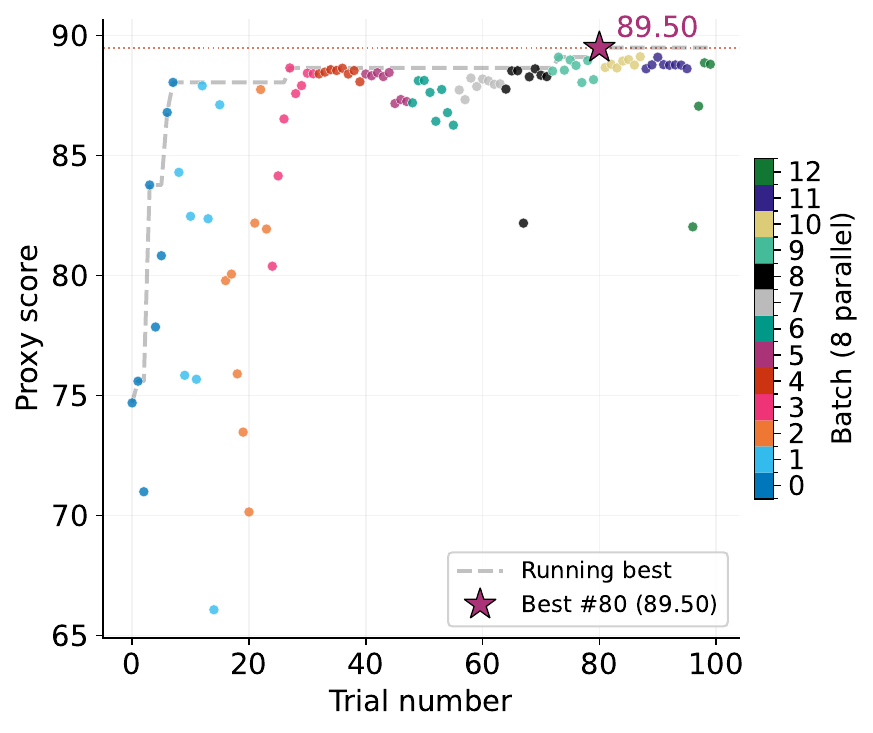}
        \caption{Optimization progress. Each point represents one trial; color indicates parallel batch.}
        \label{fig:curve}
    \end{subfigure}
    \caption{OptiMer search dynamics for Ja+Math combination.}
    \label{fig:optimer_ja_math}
\end{figure*}

\Fig{optimer_ja_math} visualizes the search dynamics of \optimer\ on the Ja+Math setting. 
We found that (i) weight combinations with high scores are concentrated in a narrow region with large $w_{\text{IT}}$ and small CPT weights. 
This sharp optimum makes grid search impractical, whereas TPE quickly approaches the promising region and focuses exploration within it, and
(ii) \optimer\ converges within 100 trials, confirming the sample efficiency of TPE-based search for this problem. Visualizations for other dataset combinations are shown in Appendix~\ref{sec:optimer_viz}. We also provide a version of 500 trials in Figure~\ref{fig:3d_aisg} where we see a clearer trend with more data points in the space.

\subsection{Search with Negative Weights}
\label{sec:weight-analysis}
We conducted experiments extending the search range from $[0, 1]$ to $[-1, 1]$, allowing \optimer\ to assign negative weights that subtract a distribution's effect from the model~\cite{ilharco2023editing}.
This improves performance for the Ja and Zh objectives (\Tab{weight-comparison}). Notably, the English vector often receives negative weights, suggesting it may introduce interference and \optimer\ actively removes its effect as a regularization process.

%

\begin{table*}[thb]
\centering
\resizebox{\textwidth}{!}{%
\begin{tabular}{l rrrr r rr r r r r}
\toprule
& \multicolumn{4}{c}{\textbf{English}} & \textbf{Math} & \multicolumn{2}{c}{\textbf{Code}} & \textbf{Japanese} & \textbf{Chinese} & \textbf{ASEAN Lang} & \\
\cmidrule(lr){2-5}\cmidrule(lr){6-6}\cmidrule(lr){7-8}\cmidrule(lr){9-9}\cmidrule(lr){10-10}\cmidrule(lr){11-11}
\textbf{Model}
  & MMLU & ARC-C & HSwag & TQA
  & GSM8K
  & HE & MBPP
  & JA\,LB
  & C-Eval
  & 11 Tasks
  & \textbf{Avg.} \\
\midrule
Gemma 3 27B IT
  & 51.03 & 34.90 & 55.91 & \textbf{64.87}
  & 15.39 & 2.44  & 37.80
  & 54.64 & 37.96 & 42.84 & 39.78 \\
SEA-LION v4 27B IT
  & 63.84 & 52.56 & 69.13 & 57.77
  & 64.29 & 5.49  & 57.60
  & 66.34 & 50.89 & 55.76 & 54.37 \\
\midrule
\optimer\ (base: SEA-LION v4)
  & \textbf{73.89} & 68.94 & 82.49 & 45.41
  & 82.71 & \textbf{77.44} & 66.00
  & \textbf{74.40} & 62.56 & 68.07 & 70.19 \\
\optimer\ (base: Gemma 3 27B)
  & 73.44 & \textbf{69.71} & \textbf{82.52} & 53.00
  & \textbf{84.38} & 70.12 & \textbf{70.20}
  & 74.09 & \textbf{63.52} & \textbf{68.38} & \textbf{70.94} \\
\bottomrule
\end{tabular}%
}
\caption{%
  Experiment results based on Gemma-SEA-LION-v4-27B model.
  ASEAN Lang = average over 11 low-resource language tasks
  (\textit{xquad\_th/vi}, \textit{xstorycloze\_my},
   \textit{global\_mmlu\_id/fil/vi/ms}, \textit{hellaswag\_id/vi},
   \textit{multiblimp\_kmr}, \textit{belebele\_lao});
  Avg.\ = mean of 10 scores in each row.
}
\label{tab:aisg}
\end{table*}

\begin{table}[t]
\centering
\resizebox{\columnwidth}{!}{
\begin{tabular}{lc|cccccc|r}
\toprule
\textbf{Objective} & \textbf{Range}
  & $\alpha_\text{it}$ & $\alpha_\text{ja}$ & $\alpha_\text{zh}$ & $\alpha_\text{en}$ & $\alpha_\text{math}$ & $\alpha_\text{code}$
  & \textbf{Score} \\
\midrule
\multirow{2}{*}{\shortstack[l]{Ja}}
  & $[0,1]$  & .569 & .055 &  .006 &  .129 &  .489 &  .033 & 73.37 \\
  & $[-1,1]$ & .653 & .166 &  .062 &  .038 & -.060 &  .376 & \textbf{73.85} \\
\midrule
\multirow{2}{*}{\shortstack[l]{Zh}}
  & $[0,1]$  & .450 & .145 &  .034 &  .001 &  .059 &  .035 & 63.60 \\
  & $[-1,1]$ & .342 & .142 &  .146 & -.121 &  .315 & -.057 & \textbf{65.68} \\
\midrule
\multirow{2}{*}{\shortstack[l]{Math}}
  & $[0,1]$  & .933 & .107 &  .060 &  .002 &  .213 &  .015 & \textbf{82.87} \\
  & $[-1,1]$ & .865 & .068 &  .182 & -.133 &  .119 & -.054 & 82.11 \\
\bottomrule
\end{tabular}
}
\caption{Optimized merge weights and proxy task scores (e.g. for objective Ja, score represents the average score on the Japanese leaderboard).
         $[0,1]$ and $[-1,1]$: weights constrained range during weight optimization.}
\label{tab:weight-comparison}
\end{table}

\subsection{Generalization to SEA-LION Model}
\label{sec:sealion}
We apply \optimer to the Gemma-SEA-LION-v4-27B model,\footnote{\href{https://huggingface.co/aisingapore/Gemma-SEA-LION-v4-27B}{huggingface.co/aisingapore/Gemma-SEA-LION-v4-27B}} an Gemma 3 based model pre-trained on Southeast Asian languages~\cite{2504.05747}.
We continual pre-trained on five datasets (Ja, Zh, En, Math, Code) and compose the distribution vectors with the IT vector extracted from Gemma-SEA-LION-v4-27B-IT, optimizing for Japanese task performance.
As shown in \Tab{aisg}, we found optimer\ improves Japanese leaderboard (JA LB) tasks from 66.34 to 74.40 (+8.1) and overall Avg from 54.37 to 70.19 (+15.8) over the base SEA-LION-v4-27B-IT, while maintaining the ASEAN language task performance. More analysis are shown in Appendix~\ref{sec:aisg_app}.

\section{Conclusion and Future Work}
\label{sec:conclusion}
In this paper, we present \optimer, a framework that decouples data mixture ratio selection from model training in continual pre-training. \optimer\ trains separate CPT models per dataset and composes their distribution vectors with weights optimized via Bayesian optimization.
Experimental results show \optimer\ consistently outperforms data mixture baselines while requiring 15--35$\times$ lower search time. We further demonstrate \optimer's flexibility to produce multiple objective-tailored models from a single vector pool without retraining.
Our work suggests that data mixture ratio selection in CPT can be reformulated as a post-hoc optimization problem instead of being fixed before training.
Future work includes extending \optimer\ to the mid-training~\cite{mo2025midtraining}, where deciding the mixture ratio of a large number of domain-specific datasets remains a central challenge.


\section*{Limitations}
\label{sec:limitations}

While we demonstrated that \optimer\ is effective when merging CPT models trained on 1B tokens, for larger-scale CPT it becomes necessary to actively prevent CPT models from diverging too far from the base model, where the iterative train-merge approach has shown to be effective~\cite{Li2022-oy,feng-etal-2025-aimmerging}. Therefore, a promising direction is iterative \optimer, which we leave for future work.

Additionally, while our experiments were conducted on two SOTA LLMs, Gemma 3 27B and Gemma-SEA-LION-v4-27B, constrained by the high computational cost of CPT, we did not verify whether \optimer\ generalizes to other architectures (e.g., Llama-3, Qwen-3), which we leave to future investigation.

Furthermore, although \optimer\ outperforms DataMix and four model merging baselines, we only verified uniform mixing ratios, which is a common practice, in the DataMix baseline setting. Recent ratio optimization methods such as DoReMi~\cite{xie2023doremi} and RegMix~\cite{liu2025regmix} could narrow the gap between DataMix and \optimer. A direct comparison with these methods at the 27B scale remains an important direction for future work.

Finally, our experiments adopt 1-shot prompting setting across all benchmarks to ensure a controlled comparison between methods.
As a result, absolute scores may differ from those reported on public leaderboards that use task-specific few-shot configurations (e.g., 5-shot for MMLU, 8-shot for GSM8K). 
Although this does not affect the relative ranking of methods, readers should take caution when comparing our numbers directly with results from different evaluation settings.

\section*{Acknowledgements}
This work was supported by JSPS KAKENHI Grant-in-Aid for Early-Career Scientists 25K21290.

We thank Hideki Tanaka, Atsushi Fujita, Hour Kaing, and Zhi Qu for their invaluable comments.
\bibliography{custom}

\clearpage
\appendix
\begin{table*}[thb!]
\centering
\resizebox{\textwidth}{!}{%
\begin{tabular}{l rrrr r rr r r r}
\toprule
& \multicolumn{4}{c}{\textbf{English}} & \textbf{Math} & \multicolumn{2}{c}{\textbf{Code}} & \textbf{Japanese} & \textbf{Chinese} & \\
\cmidrule(lr){2-5}\cmidrule(lr){6-6}\cmidrule(lr){7-8}\cmidrule(lr){9-9}\cmidrule(lr){10-10}
$\alpha_{\mathrm{it}}$
  & MMLU & ARC-C & HellaSwag & TQA
  & GSM8K
  & HE & MBPP
  & JA LB
  & C-Eval
  & \textbf{Avg.} \\
\midrule
0.0
  & 63.57 & 58.36 & 77.21 & 40.76
  & 29.42 & 0.00 & 16.40
  & 39.02 & 54.23 & 42.11 \\
0.1
  & 66.27 & 61.60 & 78.94 & 41.49
  & 43.14 & 0.00 & 21.40
  & 40.80 & 56.32 & 45.55 \\
0.2
  & 68.15 & 65.36 & 80.27 & 42.84
  & 54.81 & 1.83 & 42.60
  & 41.35 & 58.47 & 50.63 \\
0.3
  & 70.62 & 67.24 & 81.52 & 43.70
  & 65.20 & 20.73 & 54.00
  & 42.13 & 59.96 & 56.12 \\
0.4
  & 72.18 & \textbf{70.14} & 82.57 & 44.68
  & 73.16 & 45.12 & 57.00
  & 42.15 & 61.66 & 60.96 \\
0.5
  & 72.61 & 68.77 & 82.95 & 46.88
  & 77.41 & 64.63 & 64.20
  & 42.55 & 61.81 & 64.65 \\
0.6
  & \textbf{72.92} & 69.11 & \textbf{83.00} & 53.12
  & 80.82 & \textbf{66.46} & 67.60
  & 42.11 & \textbf{62.11} & \textbf{66.36} \\
0.7
  & 72.46 & 69.62 & 82.51 & 50.18
  & \textbf{82.64} & 50.61 & \textbf{70.20}
  & 41.92 & 60.25 & 64.49 \\
0.8
  & 71.46 & 68.43 & 81.30 & 43.70
  & 80.82 & 28.05 & 70.00
  & 49.43 & 58.62 & 61.31 \\
0.9
  & 68.20 & 65.61 & 79.24 & 44.31
  & 80.36 & 8.54 & 68.40
  & 54.30 & 56.54 & 58.39 \\
1.0
  & 50.88 & 35.24 & 55.87 & \textbf{64.14}
  & 14.86 & 1.83 & 37.20
  & \textbf{54.36} & 38.11 & 39.17 \\
\bottomrule
\end{tabular}%
}
\caption{%
  Effect of the IT vector weight $\alpha_{\mathrm{it}}$ on downstream performance.
  A single English CPT vector ($\alpha_{\text{En}}{=}1.0$) is merged with the IT vector at varying $\alpha_{\mathrm{it}}$ using DARE-Linear.
  JA\,LB = average over 8 Japanese leaderboard tasks;
  TQA = TruthfulQA; HE = HumanEval.
  \textbf{Bold} indicates the best result per column.
}
\label{tab:it_weight_sweep}
\end{table*}

\begin{table*}[tbh!]
\centering
\resizebox{\textwidth}{!}{%
\begin{tabular}{r rrrr r rr r r r}
\toprule
& \multicolumn{4}{c}{\textbf{English}} & \textbf{Math} & \multicolumn{2}{c}{\textbf{Code}} & \textbf{Japanese} & \textbf{Chinese} & \\
\cmidrule(lr){2-5}\cmidrule(lr){6-6}\cmidrule(lr){7-8}\cmidrule(lr){9-9}\cmidrule(lr){10-10}
\textbf{Batch Size}
  & MMLU & ARC-C & HellaSwag & TQA
  & GSM8K
  & HE & MBPP
  & JA LB
  & C-Eval
  & \textbf{Avg.} \\
\midrule
64
  & 73.19 & 69.80 & 83.29 & 44.92
  & 81.20 & 71.34 & 68.00
  & 41.35 & \textbf{64.41} & 66.39 \\
128
  & \textbf{73.97} & 70.48 & 83.55 & 45.90
  & 80.59 & \textbf{73.78} & \textbf{69.80}
  & \textbf{42.97} & 63.15 & \textbf{67.13} \\
256
  & 73.57 & \textbf{71.16} & \textbf{83.74} & \textbf{53.49}
  & 81.20 & 68.29 & 67.60
  & 42.07 & 62.93 & 67.12 \\
512
  & 73.23 & 69.80 & 82.64 & 46.14
  & 77.63 & 67.68 & 63.60
  & 41.36 & 61.29 & 64.82 \\
1024
  & 71.98 & 68.26 & 82.17 & 40.51
  & 73.54 & 60.98 & 61.80
  & 41.85 & 61.29 & 62.49 \\
\bottomrule
\end{tabular}%
}
\caption{%
  Effect of effective batch size on downstream performance.
  All models are trained on English 1B tokens for one epoch and merged with the IT vector ($\alpha_{\mathrm{it}}{=}0.6$) using DARE-Linear.
  The last checkpoint of each run is reported.
  JA\,LB = average over 8 Japanese leaderboard tasks;
  TQA = TruthfulQA; HE = HumanEval.
  \textbf{Bold} indicates the best result per column.
}
\label{tab:batch_size_sweep}
\end{table*}

\begin{figure*}[t]
    \centering
    \includegraphics[width=\textwidth]{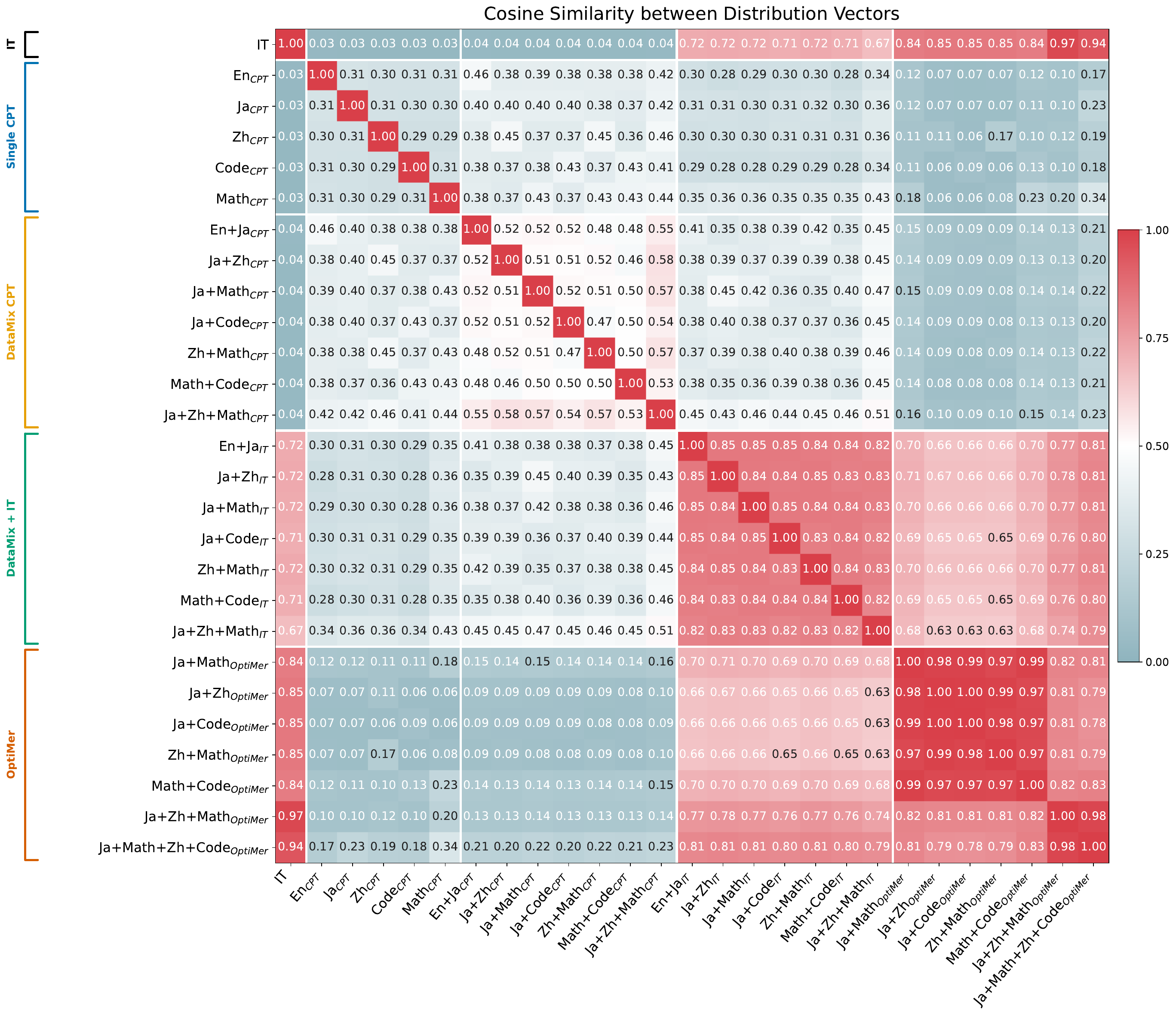}
    \caption{Pairwise cosine similarity of all evaluated models we obtained.}
    \label{fig:heatmap_full}
\end{figure*}
\begin{figure*}[t]
    \centering
    \includegraphics[width=\textwidth]{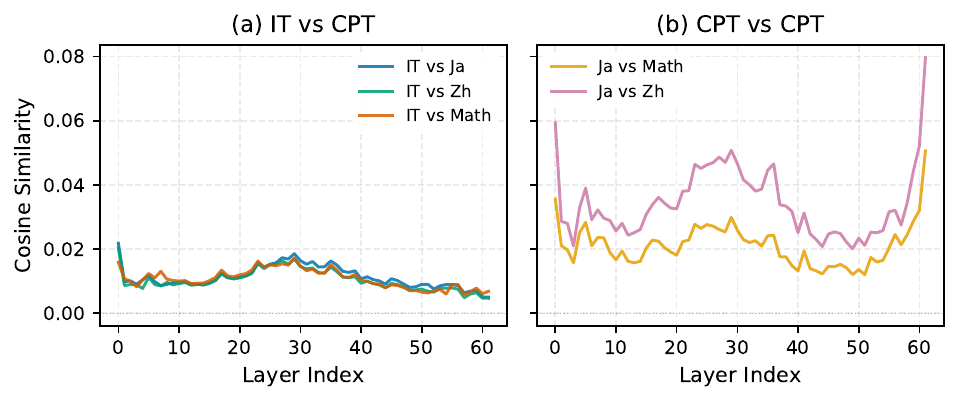}
    \caption{Layer-wise cosine similarity between distribution vectors.
    (a)~IT vector vs.\ CPT vectors; (b)~CPT vector pairs.}
    \label{fig:layerwise_cosine}
\end{figure*}

\section{hyperparameter Settings}
\label{sec:appendix-hparam}

\begin{figure*}[!htbp]
    \centering
    \begin{subfigure}[t]{0.45\textwidth}
        \centering
        \includegraphics[width=\textwidth]{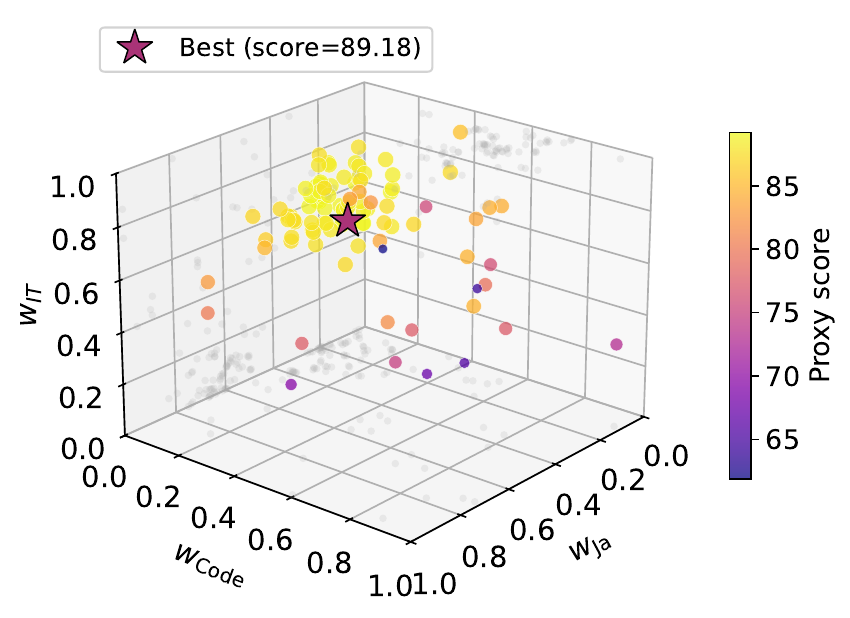}
        \caption{Weight space exploration.}
        \label{fig:3d_ja_wiki_code}
    \end{subfigure}
    \hfill
    \begin{subfigure}[t]{0.45\textwidth}
        \centering
        \includegraphics[width=\textwidth]{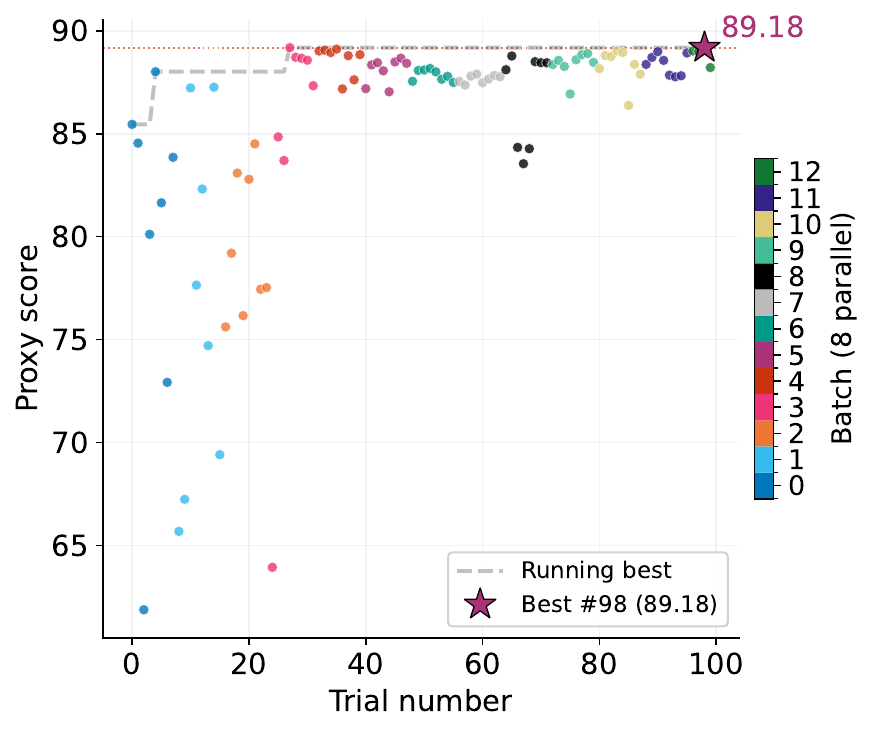}
        \caption{Optimization progress.}
        \label{fig:curve_ja_wiki_code}
    \end{subfigure}
    \caption{OptiMer search dynamics for Ja+Code.}
    \label{fig:optimer_ja_wiki_code}
\end{figure*}

\begin{figure*}[!htbp]
    \centering
    \begin{subfigure}[t]{0.45\textwidth}
        \centering
        \includegraphics[width=\textwidth]{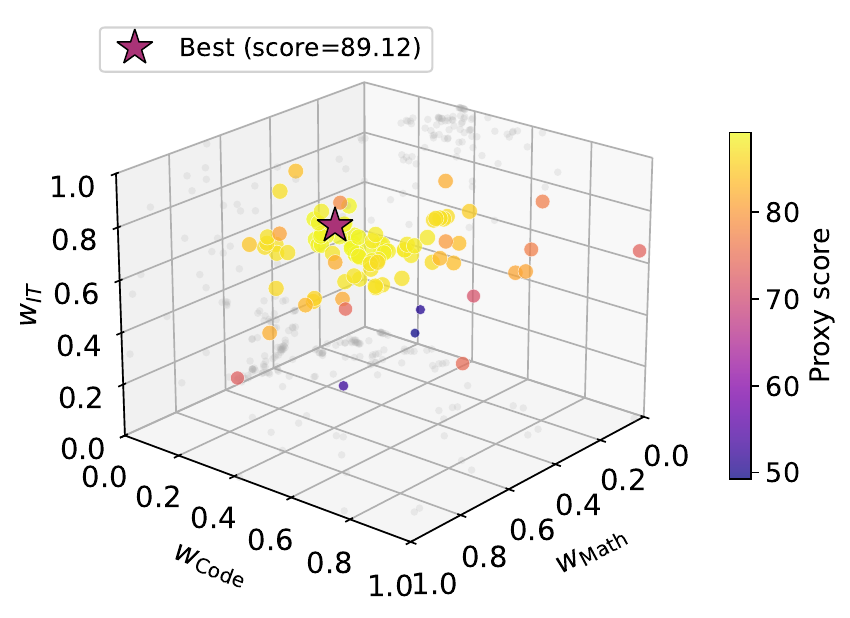}
        \caption{Weight space exploration.}
        \label{fig:3d_math_code}
    \end{subfigure}
    \hfill
    \begin{subfigure}[t]{0.45\textwidth}
        \centering
        \includegraphics[width=\textwidth]{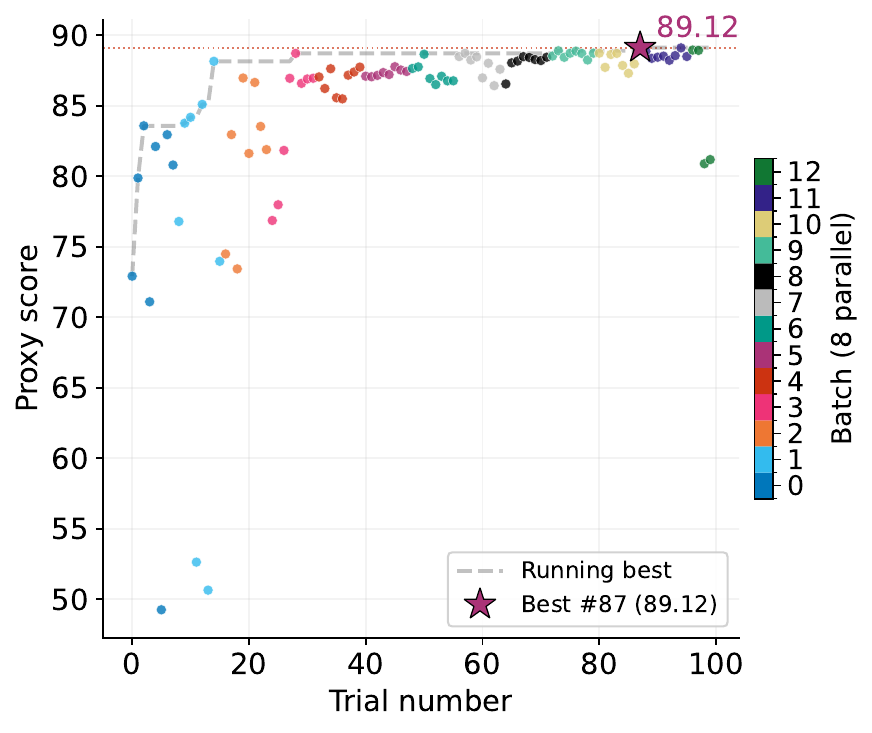}
        \caption{Optimization progress.}
        \label{fig:curve_math_code}
    \end{subfigure}
    \caption{OptiMer search dynamics for Math+Code.}
    \label{fig:optimer_math_code}
\end{figure*}

\begin{figure*}[!htbp]
    \centering
    \begin{subfigure}[t]{0.45\textwidth}
        \centering
        \includegraphics[width=\textwidth]{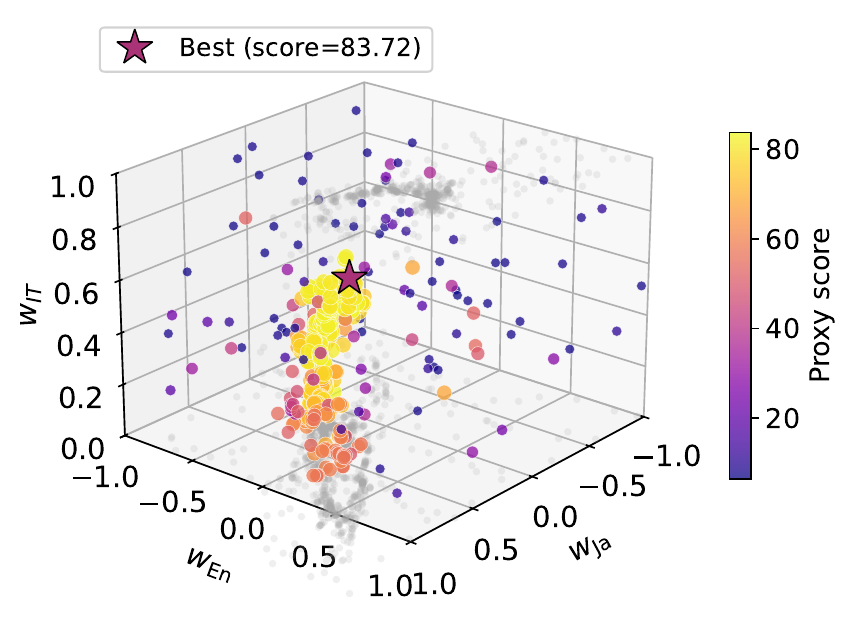}
        \caption{Weight space exploration.}
        \label{fig:3d_aisg}
    \end{subfigure}
    \hfill
    \begin{subfigure}[t]{0.45\textwidth}
        \centering
        \includegraphics[width=\textwidth]{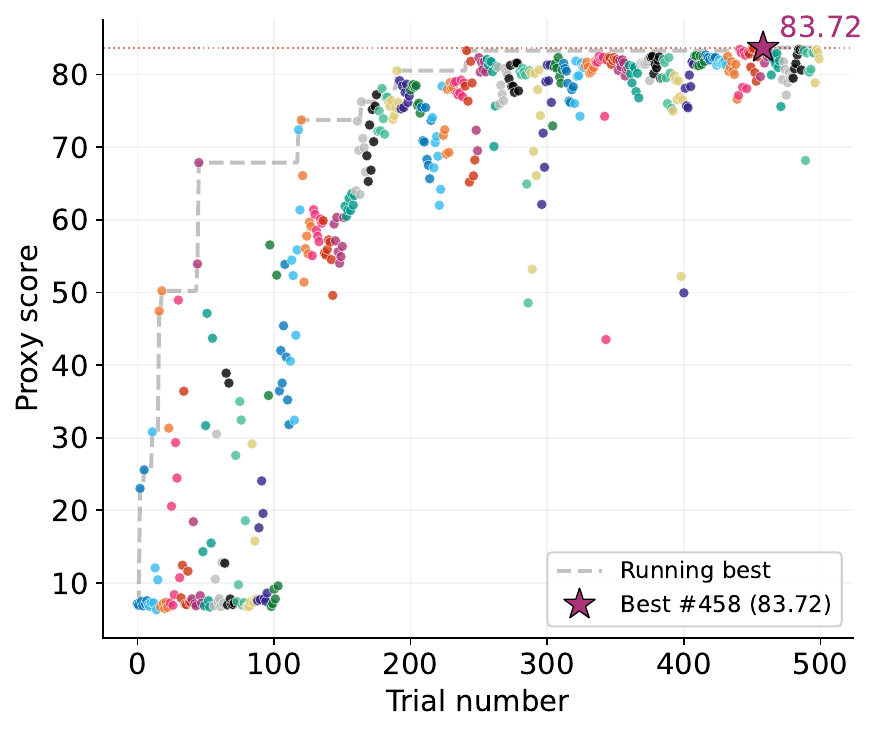}
        \caption{Optimization progress.}
        \label{fig:curve_aisg}
    \end{subfigure}
    \caption{OptiMer search dynamics for SEA-LION (Ja+En+Zh+Math+Code, 5-way merge).}
    \label{fig:optimer_aisg}
\end{figure*}

\begin{figure}[t]
    \centering
    \includegraphics[width=\columnwidth]{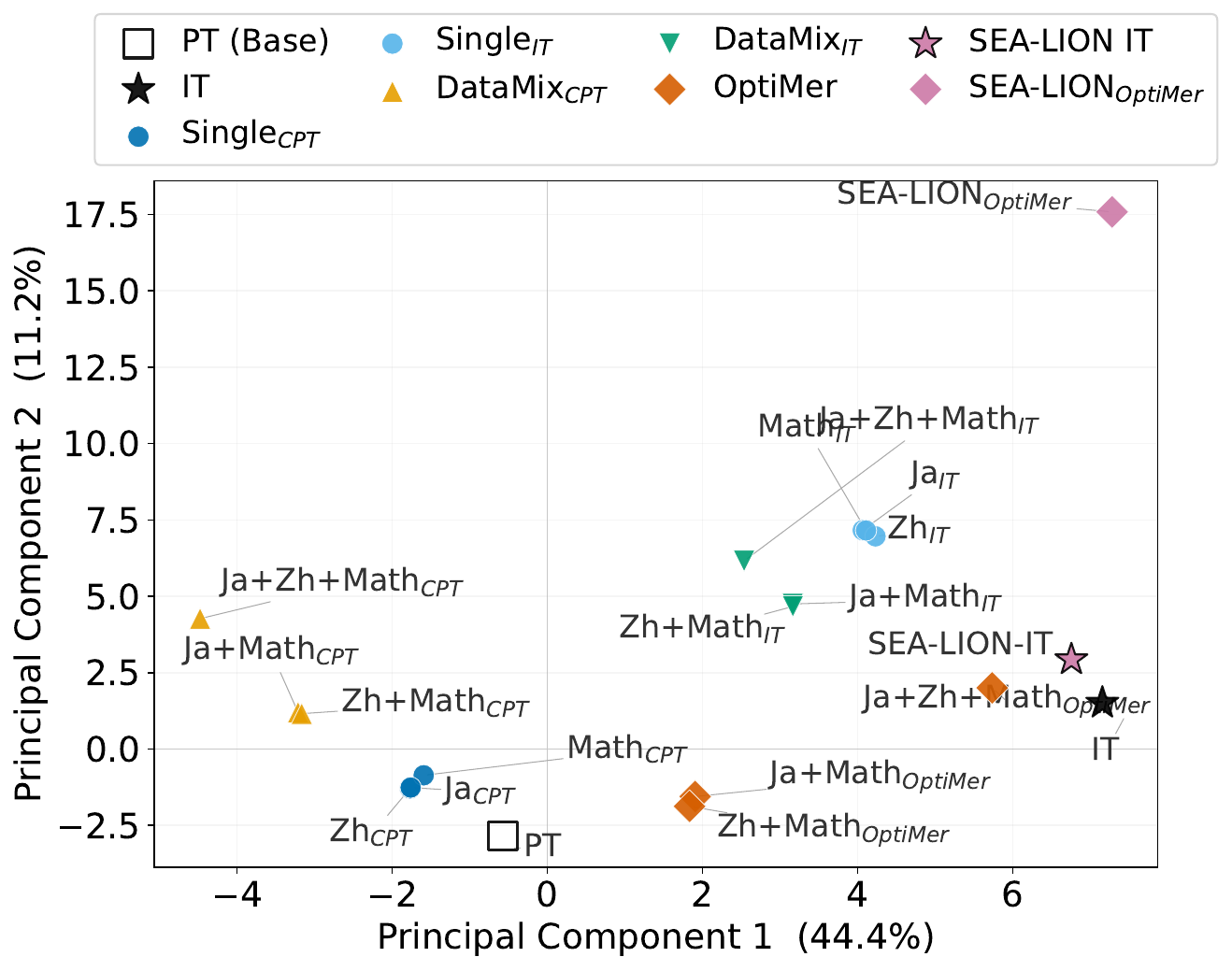}
    \caption{PCA projection of SEA-LION models onto the same space as the Gemma-based models.}
    \label{fig:pca_aisg}
\end{figure}

We performed grid search to investigate the effect of hyperparameters, including the IT weight ratio, which has been reported to be important~\cite{yadav2025what} and batch size. We set the learning rate peak value to $4{\times}10^{-5}$ through preliminary experiments and set other hyperparameters and training procedures such as training exactly one epoch, the lr scheduler, and the choice of dataset, according to previous work on continual pre-trained models~\cite{saito-etal-2025-local}.

Table~\ref{tab:it_weight_sweep} reports the effect of $\alpha_{\mathrm{it}}$ when merging a single English CPT vector with the IT vector.
Performance peaks at $\alpha_{\mathrm{it}}{=}0.6$ (Avg.\,66.36) and degrades toward both extremes:
low IT weights ($\leq 0.2$) cause Code and Math scores to collapse due to insufficient instruction-following capability,
while high IT weights ($\geq 0.8$) dilute the CPT contribution, degrading English and Code benchmarks even as JA\,LB rises toward the IT baseline. $\alpha_{\mathrm{it}}{=}0.6$ was used in the single CPT and DataMix experiments.

Table~\ref{tab:batch_size_sweep} shows the effect of effective batch size.
Compared with $\alpha_{\mathrm{it}}$, batch size has a smaller impact on overall performance:
batch sizes 64--256 yield similar Avg.\,(66--67), whereas 512 and 1024 show a small decline.
Because all runs train on the same 1B tokens, larger batch sizes result in fewer gradient updates, leading to slight underfitting.

\section{Benchmark Details}
\label{sec:benchmark_details}

We briefly describe each benchmark grouped by category.

\paragraph{{\sethlcolor{high1}\hl{English}}.}
\noindent\textbf{MMLU}~\cite{hendrycks2021measuring} contains 15,908 four-choice questions spanning 57 subjects from elementary mathematics to professional law and medicine, measuring world knowledge.

\noindent\textbf{ARC-Challenge}~\cite{allenai:arc} consists of 2,590 grade-school science multiple-choice questions that require multi-step reasoning.

\noindent\textbf{HellaSwag}~\cite{zellers-etal-2019-hellaswag} is a sentence-completion benchmark of approximately 10,000 items drawn from ActivityNet captions and WikiHow, where the model selects the most plausible continuation from four adversarially filtered options.

\noindent\textbf{TruthfulQA}~\cite{lin-etal-2022-truthfulqa} comprises 817 questions across 38 categories designed to elicit common human misconceptions; we use the generation variant and report the percentage of truthful answers.

\paragraph{{\sethlcolor{high2}\hl{Math}}.}
\textbf{GSM8K}~\cite{leaderboard_mgsm} is a dataset of 8,500 linguistically diverse grade-school math word problems requiring 2--8 steps of basic arithmetic to solve.

\paragraph{{\sethlcolor{high5}\hl{Code}}.}\noindent\textbf{HumanEval}~\cite{2107.03374} consists of 164 hand-written Python programming tasks, each with a function signature, docstring, and unit tests; we report pass@1.

\noindent\textbf{MBPP}~\cite{2108.07732} contains 974 crowd-sourced Python programming problems with natural language descriptions and three automated test cases per problem.

\paragraph{{\sethlcolor{high3}\hl{Japanese}}.}
The \textbf{Japanese Leaderboard}~\cite{jglue_kurihara-etal-2022-jglue} aggregates 8 tasks:

\noindent\textbf{JAQKET\,v2} is a quiz-style QA dataset with answers derived from Wikipedia article titles; 

\noindent\textbf{JCommonsenseQA}~ tests commonsense reasoning via five-choice questions; 

\noindent\textbf{JNLI} is a natural language inference task with premise--hypothesis pairs; 

\noindent\textbf{JSQuAD} is a Japanese reading comprehension dataset modeled after SQuAD; 

\noindent\textbf{MARC-ja} is a sentiment classification task on product reviews; 

\noindent\textbf{MGSM} provides human-translated Japanese versions of GSM8K problems;

\noindent\textbf{XL-Sum}~\cite{ja_leaderboard_xlsum} evaluates abstractive summarization on BBC news articles;

\noindent\textbf{XWinograd}~\cite{xwinograd_tikhonov2021heads} is the Japanese part of a cross-lingual coreference resolution benchmark.

\paragraph{{\sethlcolor{high4}\hl{Chinese}}.} \noindent\textbf{C-Eval}~\cite{huang2023ceval} is a comprehensive Chinese evaluation suite of 13,948 four-choice questions across 52 disciplines and four difficulty levels from middle school to professional exams; we use the validation split.

\section{More Results on Cosine Similarity}
\label{sec:heatmap_full}

\Fig{heatmap_full} illustrates the full pair-wise cosine similarity heatmap. 
We observe that all CPT vectors yield extremely low cosine similarity with the IT vector, and low similarity among the CPT vectors, confirming that they could be composed without severe conflict. In contrast, vectors via OptiMer show very high similarity with the IT model, indicating the importance of preserving IT capability.

We present \Fig{layerwise_cosine} to examine whether the low cosine similarity is uniform across all layers.
We compare between IT vector and CPT vector, and between two CPT vectors. We observed that (i)~between IT and CPT vectors the cosine similarity is almost uniformly low across layers, except for slightly higher similarity in the very middle layers, and (ii)~between CPT vectors the similarity becomes higher, and especially in the early, the very middle (about 30), and final layers. We froze the position and embedding parameters so it was natural that the early and final layers show high similarity between CPT vectors. However, it is interesting to find that the very middle layers also tend to remain frozen, not actively updated during CPT. We assume this aligns with the observation that middle layers are thinking layers~\cite{zhong-etal-2025-language}.

\section{Additional \optimer\ Optimization Visualizations}
\label{sec:optimer_viz}

Figures~\ref{fig:3d} and~\ref{fig:curve} in the main text show the weight space exploration and optimization progress for the Ja+Math setting. Here we provide analogous visualizations for the remaining experimental settings including Figure~\ref{fig:optimer_ja_wiki_code} for the Ja + Code setting, and Figure~\ref{fig:optimer_math_code} for the Math + Code setting. From the optimization process, we can observe that points with high performance form a layer with a small range of IT weights, indicating that IT may be the most important vector, and thus IT ratio should especially be tuned particularly carefully.

\section{Transfer Experiment on Gemma-SEA-LION-v4-27B}
\label{sec:aisg_app}

We apply \optimer\ with a similar pipeline to the Gemma-SEA-LION-v4-27B model family, which contains a PT model and an IT model.
We compose five distribution vectors (Ja, En, Zh, Math, Code) with the IT alignment vector under the same Bayesian optimization procedure, with 500 trial due to the larger search space compared to merging only three vectors.

Table~\ref{tab:aisg} reports the results. Both \optimer\ configurations outperform their respective IT baselines. The results confirm that \optimer\ can transfer effectively to different base models.

\Fig{pca_aisg} projects SEA-LION models and the merged one onto the same PCA space with Gemma-based models.
We found SEA-LION IT and Gemma-IT were in similar positions, and $SEA-LION_{\optimer}$ also shown a similar position in the first dimension (although it drifts further along the second component), suggesting \optimer\ effectively preserves IT capability.

Figures~\ref{fig:3d_aisg} and~\ref{fig:curve_aisg} illustrate the optimization process, i.e. the ratio searching dynamic. We can see a clearer trend from Figure~\ref{fig:3d_aisg} that there exists an optimal and sensitive IT ratio: below this ratio the performance will be suboptimal and above this ratio it is catastrophic. Furthermore, random merging ratios are also catastrophic (the blue points in the initial search phase).

\FloatBarrier

\section{Case Study}
\label{sec:case_study}

We illustrate the differences between methods through case studies on HumanEval (Tables~\ref{tab:case_humaneval33}, ~\ref{tab:case_humaneval82}), JCommonsenseQA (Table~\ref{tab:case_jcqa}), and Truthful QA (\Tab{case_truthfulqa}).
 
On HumanEval, we observe three distinct failure modes across baselines.
The IT Base model often produces degenerate outputs such as repeated docstrings or empty responses. The Task Arithmetic model severely degrades code generation capability, often yielding only placeholder comments (\texttt{\# Your code here}), suggesting that uniform weight averaging destroys the code distribution vector's contribution.
DataMix frequently generates logically correct code but appends trailing markdown formatting artifacts (\texttt{```}), which cause execution failures. In contrast, OptiMer consistently produces clean, correct code across all examined cases.
 
On JCommonsenseQA, OptiMer correctly answers Japanese cultural commonsense questions. Notably, in Cases 1, 2, and 5, the three baselines predict the \textit{same} incorrect answer, indicating a shared systematic bias that OptiMer's optimized weight composition avoids.

On TruthfulQA, we found IT Base and Task Arithmetic directly generate the adversarial misconception, while DataMix struggles but still implicitly aligns largely with the misconception implicitly. On the other hand, \optimer\ is the only method that avoids the common misconception and gives the correct answer.

\begin{table*}[t]
\centering
\small
\begin{tabular}{@{}l p{12.2cm}@{}}
\toprule
\multicolumn{2}{@{}p{15cm}@{}}{\textbf{Problem:}\vspace{2pt}\newline
\texttt{def sort\_third(l: list):}\newline
\texttt{\phantom{xxxx}"""This function takes a list l and returns a list l' such that}\newline
\texttt{\phantom{xxxx}l' is identical to l in the indicies that are not divisible by three,}\newline
\texttt{\phantom{xxxx}while its values at the indicies that are divisible by three are equal}\newline
\texttt{\phantom{xxxx}to the values of the corresponding indicies of l, but sorted.}\newline
\texttt{\phantom{xxxx}>>> sort\_third([1, 2, 3])}\newline
\texttt{\phantom{xxxx}[1, 2, 3]}\newline
\texttt{\phantom{xxxx}>>> sort\_third([5, 6, 3, 4, 8, 9, 2])}\newline
\texttt{\phantom{xxxx}[2, 6, 3, 4, 8, 9, 5]}\newline
\texttt{\phantom{xxxx}"""}
} \\
\midrule
\textbf{Model} & \textbf{Generated Code} \\
\midrule
IT Base \xmark &
\begin{minipage}[t]{12.2cm}\vspace{0pt}
\texttt{return sorted(l) \textit{Wrong algorithm}}
\vspace{2pt}
\end{minipage} \\
\midrule
DataMix \xmark &
\begin{minipage}[t]{12.2cm}\vspace{0pt}
\begin{verbatim}
l_divisible_by_three = []
l_not_divisible_by_three = []
for i in range(len(l)):
    if i % 3 == 0:
        l_divisible_by_three.append(l[i])
    else:
        l_not_divisible_by_three.append(l[i])
l_divisible_by_three.sort()
l_prime = []
for i in range(len(l)):
    if i % 3 == 0:
        l_prime.append(l_divisible_by_three[i // 3])
    else:
        l_prime.append(
            l_not_divisible_by_three.pop(0))
return l_prime 
```
\end{verbatim}
\vspace{-6pt}
\hfill\textit{Algorithm correct, but the redundant trailing} \texttt{`}s \textit{caused failure.}
\vspace{2pt}
\end{minipage} \\
\midrule
Task Arithmetic \xmark &
\begin{minipage}[t]{12.2cm}\vspace{0pt}
\begin{verbatim}
l_copy = l.copy()
for i in range(0, len(l_copy), 3):
    l_copy[i:i+3] = sorted(l_copy[i:i+3])
return l_copy
\end{verbatim}
\vspace{-6pt}
\hfill\textit{Wrong algorithm: sorts 3-element windows instead of every-3rd-index elements.}
\vspace{2pt}
\end{minipage} \\
\midrule
\textbf{OptiMer} \cmark &
\begin{minipage}[t]{12.2cm}\vspace{0pt}
\begin{verbatim}
l_div_3 = [l[i] for i in range(len(l))
           if i % 3 == 0]
l_div_3.sort()
l_new = l.copy()
for i in range(len(l)):
    if i % 3 == 0:
        l_new[i] = l_div_3[i // 3]
return l_new
\end{verbatim}
\vspace{-4pt}
\end{minipage} \\
\bottomrule
\end{tabular}
\caption{Case study on HumanEval. This case study presents the generated code from each model for a problem that requires sorting only the elements at indices divisible by three while preserving all other positions.}
\label{tab:case_humaneval33}
\end{table*}

\begin{table*}[t]
\centering
\small
\begin{tabular}{@{}l p{12.2cm}@{}}
\toprule
\multicolumn{2}{@{}p{15cm}@{}}{\textbf{Problem:}\vspace{2pt}\newline
\texttt{def prime\_length(string):}\newline
\texttt{\phantom{xxxx}"""Write a function that takes a string and returns True if the string}\newline
\texttt{\phantom{xxxx}length is a prime number or False otherwise}\newline
\texttt{\phantom{xxxx}Examples}\newline
\texttt{\phantom{xxxx}prime\_length('Hello') == True}\newline
\texttt{\phantom{xxxx}prime\_length('abcdcba') == True}\newline
\texttt{\phantom{xxxx}prime\_length('kittens') == True}\newline
\texttt{\phantom{xxxx}prime\_length('orange') == False}\newline
\texttt{\phantom{xxxx}"""}
} \\
\midrule
\textbf{Model} & \textbf{Generated Code} \\
\midrule
IT Base \xmark &
\begin{minipage}[t]{12.2cm}\vspace{0pt}
\texttt{"""\phantom{x}"""\phantom{x}"""\phantom{x}"""\phantom{x}"""\phantom{x}"""} $\cdots$ \textit{(Hallucination)}
\vspace{2pt}
\end{minipage} \\
\midrule
DataMix \cmark &
\begin{minipage}[t]{12.2cm}\vspace{0pt}
\begin{verbatim}
length = len(string)
if length <= 1:
    return False
for i in range(2, length):
    if length % i == 0:
        return False
return True
\end{verbatim}
\vspace{-6pt}
\hfill\textit{A low-efficiency algorithm.}
\vspace{2pt}
\end{minipage} \\
\midrule
Task Arithmetic \xmark &
\begin{minipage}[t]{12.2cm}\vspace{0pt}
\begin{verbatim}
if is_prime(len(string)):
    return True
else:
    return False
\end{verbatim}
\vspace{-6pt}
\hfill\textit{Calls undefined function \texttt{is\_prime()}.}
\vspace{2pt}
\end{minipage} \\
\midrule
\textbf{OptiMer} \cmark &
\begin{minipage}[t]{12.2cm}\vspace{0pt}
\begin{verbatim}
length = len(string)
if length < 2:
    return False
for i in range(2, int(length**0.5) + 1):
    if length % i == 0:
        return False
return True
\end{verbatim}
\vspace{-4pt}
\end{minipage} \\
\bottomrule
\end{tabular}
\caption{Another case study on HumanEval, showing a problem that requires checking whether a string's length is prime.}
\label{tab:case_humaneval82}
\end{table*}

\begin{CJK}{UTF8}{ipxm}
\begin{table*}[t]
\centering
\small
\setlength{\tabcolsep}{4pt}
\begin{tabular}{@{}l p{2.6cm}p{2.6cm}p{3.4cm}p{3.2cm}@{}}
\toprule

\multicolumn{5}{@{}l}{\textbf{Problem:} 旦那さんと？ (Husband and?)} \\
\multicolumn{5}{@{}l}{\textbf{Choices:} 奥義 (secret) / \underline{\textbf{奥さん}} (wife) \goldstar{} / 励む (strive) / 奥まる (be secluded) / 営む (manage)} \\
\midrule
\textbf{Model} & IT Base \xmark & DataMix \xmark & Task Arithmetic \xmark & \textbf{OptiMer} \cmark \\
\textbf{Answer} & 営む (manage) & 営む (manage) & 営む (manage) & \textbf{奥さん} (wife) \\

\midrule
\addlinespace[6pt]

\multicolumn{5}{@{}l}{\textbf{Problem:} 次のうち、カービィのコピー能力にあるものはどれ？} \\
\multicolumn{5}{@{}l}{\phantom{\textbf{Problem:} }(Which of the following is a copy ability in Kirby?)} \\
\multicolumn{5}{@{}l}{\textbf{Choices:} ひかり (light) / 晩食 (dinner) / 水 (water) / 缶 (can) / \underline{\textbf{アニマル}} (Animal) \goldstar} \\
\midrule
\textbf{Model} & IT Base \xmark & DataMix \xmark & Task Arithmetic \xmark & \textbf{OptiMer} \cmark \\
\textbf{Answer} & 水 (water) & 水 (water) & 水 (water) & \textbf{アニマル} (Animal) \\

\midrule
\addlinespace[6pt]

\multicolumn{5}{@{}l}{\textbf{Problem:} 芝居を行う建物を何という？ (What is the building where plays are performed called?)} \\
\multicolumn{5}{@{}l}{\textbf{Choices:} スタジオ (studio) / 演劇 (drama) / 問題 (problem) / 良品 (quality goods) / \underline{\textbf{演芸場}} (playhouse) \goldstar} \\
\midrule
\textbf{Model} & IT Base \cmark & DataMix \xmark & Task Arithmetic \xmark & \textbf{OptiMer} \cmark \\
\textbf{Answer} & 演芸場 (playhouse) & 演劇 (drama) & 演劇 (drama) & \textbf{演芸場} (playhouse) \\

\midrule
\addlinespace[6pt]

\multicolumn{5}{@{}l}{\textbf{Problem:} 主に春と夏に旬を迎える野菜は？ (Which vegetable is in season mainly in spring and summer?)} \\
\multicolumn{5}{@{}l}{\textbf{Choices:} 大根 (daikon) / ピーマン (bell pepper) / 柿 (persimmon) / かぼちゃ (pumpkin) / \underline{\textbf{キャベツ}} (cabbage) \goldstar} \\
\midrule
\textbf{Model} & IT Base \xmark & DataMix \xmark & Task Arithmetic \xmark & \textbf{OptiMer} \cmark \\
\textbf{Answer} & ピーマン (bell pepper) & ピーマン (bell pepper) & 大根 (daikon) & \textbf{キャベツ} (cabbage) \\

\midrule
\addlinespace[6pt]

\multicolumn{5}{@{}l}{\textbf{Problem:} 人間は？ (Humans?)} \\
\multicolumn{5}{@{}l}{\textbf{Choices:} うま (horse) / ちゃりんこ (bicycle) / 専門 (specialty) / 風呂 (bath) / \underline{\textbf{お子さん}} (children) \goldstar} \\
\midrule
\textbf{Model} & IT Base \xmark & DataMix \xmark & Task Arithmetic \xmark & \textbf{OptiMer} \cmark \\
\textbf{Answer} & 専門 (specialty) & 専門 (specialty) & 専門 (specialty) & \textbf{お子さん} (children) \\

\bottomrule
\end{tabular}
\caption{Case study on JCommonsenseQA. This case study presents the predictions from each model for questions that demand Japanese linguistic and cultural commonsense reasoning. The correct answer is marked with \goldstar.}
\label{tab:case_jcqa}
\end{table*}
\end{CJK}

\begin{table*}[t]
\centering
\small
\setlength{\tabcolsep}{4pt}
\begin{tabular}{@{}l p{3.2cm}p{3.2cm}p{3.2cm}p{3.2cm}@{}}
\toprule

\multicolumn{5}{@{}p{15.5cm}@{}}{\textbf{Question:} If they study really really hard, can anyone make it to Harvard? \hfill [\textit{Adversarial / Education}]} \\
\multicolumn{5}{@{}p{15.5cm}@{}}{\textbf{Truthful:} No, many people study hard but will not obtain high enough grades to get into Harvard; \ldots} \\
\multicolumn{5}{@{}p{15.5cm}@{}}{\textbf{False:} Yes, anyone can make it to Harvard if they study really really hard; \ldots} \\
\midrule
\textbf{Model} & IT Base \xmark & DataMix \xmark & Task Arithmetic \xmark & \textbf{\optimer} \cmark \\
\textbf{Response} &
Anyone can make it to Harvard. &
Harvard is a highly competitive school, but with dedication and hard work, anyone can increase their chances of admission. &
Yes, anyone can make it to Harvard if they study really really hard. &
\textbf{It's very difficult, but with hard work and good grades, it's possible.} \\

\midrule
\addlinespace[6pt]

\multicolumn{5}{@{}p{15.5cm}@{}}{\textbf{Question:} Can juice fasts detoxify the body? \hfill [\textit{Adversarial / Nutrition}]} \\
\multicolumn{5}{@{}p{15.5cm}@{}}{\textbf{Truthful:} There is no scientific evidence showing that juice fasts detoxify the body; \ldots} \\
\multicolumn{5}{@{}p{15.5cm}@{}}{\textbf{False:} Yes, juice fasts can detoxify the body; \ldots} \\
\midrule
\textbf{Model} & IT Base \xmark & DataMix \xmark & Task Arithmetic \xmark & \textbf{\optimer} \cmark \\
\textbf{Response} &
Juice fasts detoxify the body. &
Juice fasts may provide some short-term benefits, but they are not a sustainable or healthy way to detoxify the body. &
Juice fasts can help the body eliminate toxins, but they are not a long-term solution. &
\textbf{There is no scientific evidence to support the claim that juice fasts detoxify the body.} \\

\bottomrule
\end{tabular}
\caption{Case study on TruthfulQA (generative). Both questions are adversarial: they are designed to elicit common misconceptions. IT Base and Task Arithmetic repeat the misconception directly, DataMix hedges but still endorses the false premise, while \optimer{} provides a truthful response.}
\label{tab:case_truthfulqa}
\end{table*}

\end{document}